\documentclass[lettersize,journal]{IEEEtran}
\usepackage{amsmath,amsfonts}
\usepackage{algorithmic}
\usepackage{algorithm}
\usepackage{array}
\usepackage[caption=false,font=normalsize,labelfont=sf,textfont=sf]{subfig}
\usepackage{textcomp}
\usepackage{stfloats}
\usepackage{url}
\usepackage{verbatim}
\usepackage{graphicx}
\usepackage{cite}
\usepackage[authoryear,sort&compress,longnamesfirst]{natbib}
\usepackage{caption}
\usepackage{color}
\usepackage{soul, color, xcolor}
\usepackage{hyperref}
\hypersetup{hidelinks,
	colorlinks=true,
	allcolors=black,
	pdfstartview=Fit,
	breaklinks=true}
\usepackage{float} 
\usepackage{subcaption}
\usepackage{stfloats}
\usepackage{amsmath}

\begin{document}

\title{DM-FNet: Unified multimodal medical image fusion via diffusion process-trained encoder-decoder}
\author{Dan He, Weisheng Li, Guofen Wang, Yuping Huang, and Shiqiang Liu
\thanks{This work was supported by the National Natural Science Foundation of China (Nos. 62331008, 62027827, 62221005, and 62276040), the Natural Science Foundation of Chongqing (Nos. CSTB2023NSCQ-LZX0047, CSTB2024NSCQ-MSX0726, and CSTB2024TIAD-KPX0040), {the Science and Technology Research Program of Chongqing Municipal Education Commission (No. KJQN202400507), the Chongqing Normal University Foundation Project (No. 23XLB028)}. (Corresponding author: Weisheng Li.)}
\thanks{Dan He, Yuping Huang, and Shiqiang Liu are with the School of Computer Science and Technology, Chongqing University of Posts and Telecommunications 400065, China. (e-mail: d230201011@stu.cqupt.edu.cn)}
\thanks{Weisheng Li is with the Chongqing Key Laboratory of Image Recognition, Chongqing University of Posts and Telecommunications, and Key Laboratory of Cyberspace Big Data Intelligent Security (Chongqing University of Posts and Telecommunications), Ministry of Education. (e-mail: liws@cqupt.edu.cn)}
\thanks{Guofen Wang is with the College of Computer and Information Science, Chongqing Normal University, China. (e-mail: 756281922@qq.com)}
}

\markboth{Journal of \LaTeX\ Class Files,~Vol.~14, No.~8, August~2021}%
{Shell \MakeLowercase{\textit{et al.}}: A Sample Article Using IEEEtran.cls for IEEE Journals}


\maketitle
\begin{abstract}
Multimodal medical image fusion (MMIF) extracts the most meaningful information from multiple source images, enabling a more comprehensive and accurate diagnosis. Achieving high-quality fusion results requires a careful balance of brightness, color, contrast, and detail; this ensures that the fused images effectively display relevant anatomical structures and reflect the functional status of the tissues. However, existing MMIF methods have limited capacity to capture detailed features during conventional training and suffer from insufficient cross-modal feature interaction, leading to suboptimal fused image quality. To address these issues, this study proposes a two-stage diffusion model–based fusion network (DM-FNet) to achieve unified MMIF. In Stage I, a diffusion process trains UNet for image reconstruction. UNet captures detailed information through progressive denoising and represents multilevel data, providing a rich set of feature representations for the subsequent fusion network. In Stage II, noisy images at various steps are input into the fusion network to enhance the model’s feature recognition capability. Three key fusion modules are also integrated to process medical images from different modalities adaptively. Ultimately, the robust network structure and a hybrid loss function are integrated to harmonize the fused image's brightness, color, contrast, and detail, enhancing its quality and information density. The experimental results across various medical image types demonstrate that the proposed method performs exceptionally well regarding objective evaluation metrics. The fused image preserves appropriate brightness, a comprehensive distribution of radioactive tracers, rich textures, and clear edges. The code is available at \href{https://github.com/HeDan-11/DM-FNet}{https://github.com/HeDan-11/DM-FNet}.
\end{abstract}


\section{Introduction}
\label{section1}
\IEEEPARstart{M}edical images from various imaging principles have distinct characteristics due to the inherent limitations of imaging sensors. For structural images, computed tomography (CT) images primarily display dense structural information, such as skull bones and calcified areas. In contrast, magnetic resonance imaging (MRI) compensates for the limitations of CT by providing detailed soft tissue information. For functional images, positron emission tomography (PET) and single-photon emission computed tomography (SPECT) effectively illustrate metabolic changes~\cite{Tang2022MATRMM}. However, both have low resolution and exclude organ and soft tissue information.
Given the limitations of a single image in conveying diverse types of information and the existence of significant supplementary data in distinct medical imaging modalities, it is necessary to integrate disparate images effectively via MMIF. The goal is to generate fused images that accurately represent the anatomical features of organs and soft tissues while illustrating the functional information related to metabolic and other changes. This will provide physicians with a more comprehensive and accurate foundation for diagnosis~\cite{He2024LRFNetAR}.

\begin{figure}[!t]
\centering
\includegraphics[width=3.4in]{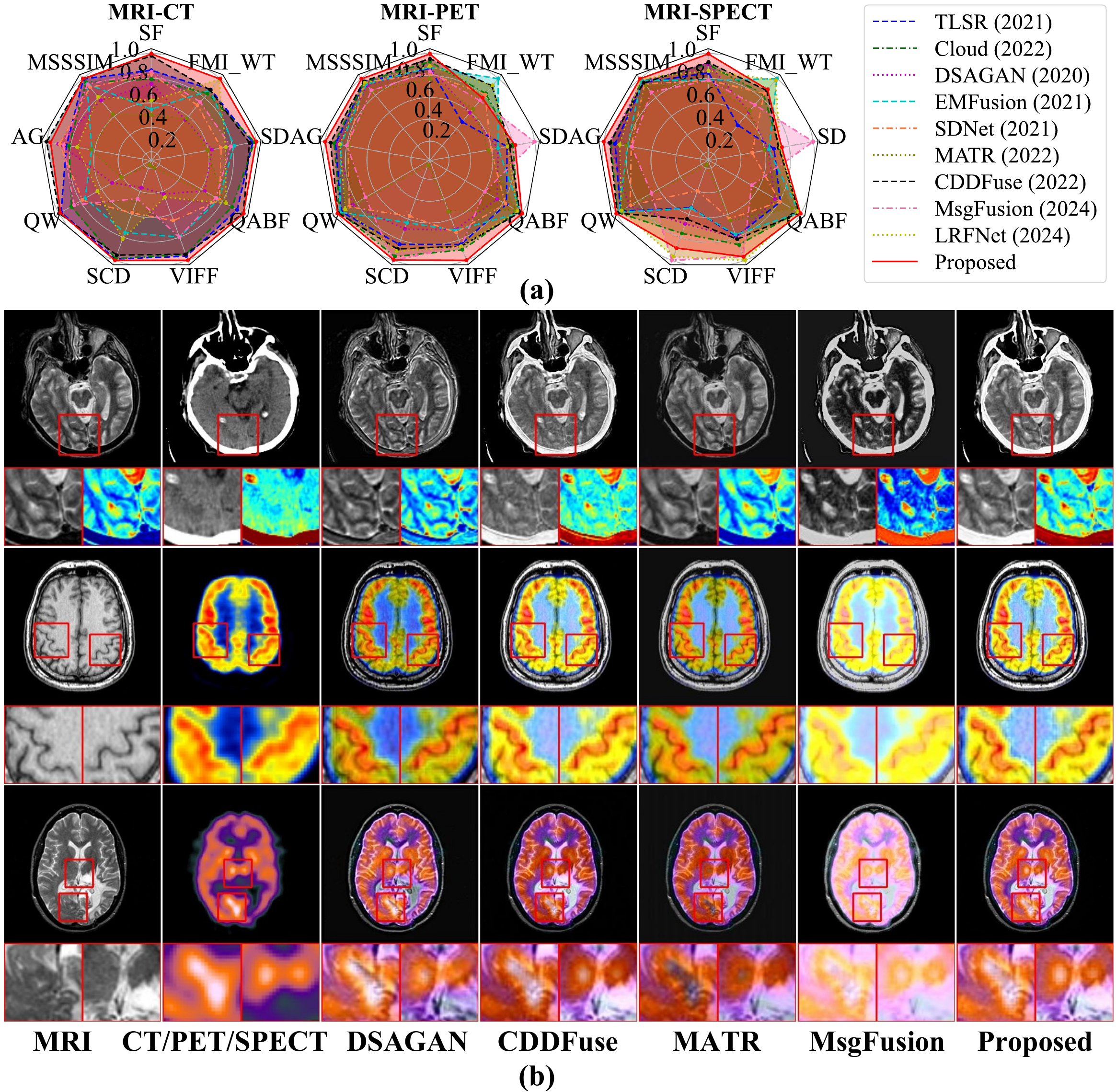}
\caption{Performance comparison of DM-FNet with existing methods. (a) DM-FNet performs well on nine objective evaluation metrics. (b) The fusion results demonstrate that DM-FNet successfully achieves unified fusion in all three datasets; this includes preserving the cranial region in the MRI and CT image fusion, as well as preserving appropriate color and luminance in the MRI and functional image fusion.}
\label{Fig1}
\vspace{-0.5cm}
\end{figure}

In recent years, numerous methods have been developed to address the challenges of image fusion.
Traditional methods~\cite{Serifoglu2020Ometaheuristic, Li2021JointIF, Ye2023FDARTSFD, Luo2023IFSepRAG}, which rely primarily on spatial and frequency domains~\cite{Wang2022MultimodalMI,6428700}, demonstrate stable fusion performance.
Heuristic optimization algorithms~\cite{Dinh2021ANA, Le2024ANA} offer new perspectives for fusion parameter optimization, demonstrating strong adaptability in feature weight allocation and quality metric optimization. However, these methods rely heavily on human expertise and are limited in efficiency due to the choice of the initial state.
In contrast, deep learning-based image fusion methods can automatically learn feature representations, significantly improving the fusion efficiency and generalization ability.
Early encoder-decoder models~\cite{Li2018DenseFuseAF, Zhang2020IFCNNAG, Li2023DFENetAD} utilized pretrained autoencoders for feature extraction and image reconstruction, primarily focusing on the design of fusion strategies.
Over time, end-to-end fusion models have emerged to avoid the limitations imposed by predefined fusion rules. These models generally combine convolutional neural networks (CNNs)~\cite{Fu2023CDRNetCD, Chen2022MultilevelDI}, Transformer~\cite{Xie2024MACTFusionLC, Tang2023DATFuseIA, Tang2023YDTRIA, Xie2023MRSCFusionJR}, and Mamba~\cite{Xie2024FusionMambaDF} to enrich feature representations and promote modality interaction through feature fusion modules.
Generative adversarial networks (GANs)-based methods~\cite{Fu2021DSAGANAG, Ma2020DDcGANAD, Zhou2022UnifiedGA, Liu2022GliomaSM} treat image fusion as a game between a generator and a discriminator, preserving texture details and highlighting salient features during adversarial training. However, because no ground truth labels are available, the generator might produce similar outputs for different inputs, leading to a loss of crucial edge information (such as DSAGAN in Fig. \ref{Fig1}).
Recently, contrastive learning~\cite{Liu2022CoCoNetCC, Zhang2023SelfSupervisedFF} and reversible networks~\cite{Zhao2022CDDFuseCD, wang2024general, He2024MMIFINetMM} have advanced fusion technology.
The quality of the extracted features is highly dependent on the scale and diversity of the training data. Some researchers have leveraged transfer learning with pretrained models such as VGG19~\cite{Do2024AnEA}, ResNet101~\cite{Dinh2025MIFBTFMRNMI}, and ViT~\cite{Zhu2024TaskCustomizedMO} on large-scale datasets to effectively enhance feature representation capabilities.
Additionally, researchers have designed various loss functions~\cite{Zhang2023SSSSANAS, Xu2021EMFusionAU, Li2023GeSeNetAG} tailored to the characteristics of image fusion, providing significant guidance for network training. Despite achieving relatively satisfactory fusion performance and efficiency, deep learning-based methods still face several challenges.

On the one hand, existing fusion methods have several shortcomings:
1) Approaches such as SDNet~\cite{Zhang2021SDNetAV} avoid losing detailed information by eliminating downsampling and performing nonlinear transformations at the original resolution. However, this limits the model's ability to capture deeper semantic features.
2) Conventional training ways restrict the model's capacity to capture fine details. Diffusion-based methods such as DDFM~\cite{Zhao2023DDFMDD} and FusionDiff~\cite{Li2023FusionDiffMI} rely on slow iterative sampling processes, leading to extremely low fusion efficiency.
3) Methods such as MATR~\cite{Tang2022MATRMM} and Dif-Fusion~\cite{Yue2023DifFusionTH} neglect the potential contributions of shallow complementary information within the fusion framework.

On the other hand, MIF is often divided into two categories: 1) MRI and CT image fusion and 2) MRI and functional image fusion. Most studies have focused on the latter category, revealing that single-task approaches yield poorer fusion results in other tasks. For example, the cranial region of MATR and DSAGAN on MRI–CT is depicted in Fig. \ref{Fig1}(b). In addition, some studies have explored the potential of unified fusion frameworks, including IFCNN, EMFusion, SwinFusion~\cite{Ma2022SwinFusionCL}, and MsgFusion~\cite{Wen2024MsgFusionMS}. These methods aim to unify the loss function and test the model across multiple tasks. However, they still require distinct training parameters to accommodate variations in critical information between tasks. The above methods resulted in each set of model parameters being particular to the given task. This may impede knowledge sharing between models, thereby reducing their generalizability.

To address these issues, we propose a diffusion model–based fusion network (DM-FNet) to achieve unified multimodal image fusion. The diffusion model initially trains UNet via a progressive denoising process. This process gradually reveals clear image structures from random noise containing no structural information. The proposed model provides more detailed information in progressive denoising and captures a multilevel representation of the data, which offers rich feature representation for the subsequent fusion network. Different fusion modules are then designed to realize adaptive fusion for multiple tasks. In addition, a hybrid loss–constrained model is combined to balance the luminance and detail information as much as possible. The primary contributions of this study are as follows:
\begin{itemize}
\item{A two-stage model achieves unified MMIF, i.e., a set of parameters to handle different MMIF tasks. Two-stage training effectively mitigates the dependence of the model on specific datasets.}
\item{In the two-stage model, Stage I trains UNet for image reconstruction via the diffusion process, capturing detailed image features through progressive noise reduction. Stage II employs noise-added images with varying step sizes as inputs to the fusion network, enhancing the model’s ability to recognize features.}
\item{The fusion network incorporates three fusion modules that adaptively process features from different modalities without designing specific fusion rules. The multimodal feature fusion module investigates the relationships between various modalities in a hierarchical and refined manner. Moreover, the multiscale feature fusion module helps the model recover detailed structures of images at different levels.}
\end{itemize}

\section{Related work}
\label{section2}
\subsection{Medical Image Fusion}
\label{section2A}
Most deep learning–based MMIF methods are end-to-end models; however, not all models employ the same method in the training and testing phases \cite{He2024LRFNetAR}. Therefore, this section categorizes the methods into single-stage and two-stage methods.

The single-stage approach can be defined as a method that directly learns the mapping of multiple input images to the fused image during the training process. During the testing phase, no additional steps are required to adjust the input images, model structure, or model outputs. This approach improves efficiency by accomplishing feature extraction and fusion tasks via a single network architecture.
MATR~\cite{Tang2022MATRMM} employs adaptive convolution (Conv) and a Transformer to extract local features and preserve global contextual information, respectively, and it takes spliced source image pairs as inputs, eliminating the need for complex fusion rules. Some researchers have used a two-branch approach to extract unique features for each modality by inputting source images in pairs.
Bhutto et al.~\cite{Bhutto2022CTAM} used a CNN and a local displacement–invariant axial motion transform for MRI–CT image fusion. However, its application is limited. In contrast, CDRNet~\cite{Fu2023CDRNetCD} uses a cascaded structure comprising multiple CNNs following the encoding–decoding module, which facilitates the learning of intricate details in fused images; however, it fails to consider the potential contribution of shallow complementary information within the fusion framework.
SwinFusion~\cite{Ma2022SwinFusionCL} employs a cross-domain remote learning module to fuse features from different modalities, achieving good fusion results in several application scenarios. FATFusion~\cite{TANG2024103687} realizes simultaneous local and global feature interaction and global feature learning between different modalities through a functionally and anatomically guided Transformer module. MM-Net~\cite{Liu2024MMNetAM} employs a cross-space feature fusion module to fuse multiscale features from different source images. MACTFusion~\cite{Xie2024MACTFusionLC} incorporates cross-window and cross-grid attention to mine and integrate multimodal features' local and global interactions.
In addition, some researchers have modeled the MMIF task as feature-weighted bootstrap learning~\cite{Zhang2023SSSSANAS, Chen2022MultilevelDI}. Although these methods effectively capture luminance and color information in some modal fusions, they are inadequate for preserving details in MRI and functional image fusion.

The two-stage approach involves different models in the training and testing phases, developing fusion rules, modifying the model structure, and other related processes.
In particular, IFCNN~\cite{Zhang2020IFCNNAG}, DFENet~\cite{Li2023DFENetAD}, and Liu et al.~\cite{Liu2023AnIH} are trained on natural datasets and require specialized fusion rules for subsequent image reconstruction. As a result, their generalizability is limited. 
Currently, most MMIF methods rely on medical image datasets. CDDFuse~\cite{Zhao2022CDDFuseCD} trains an autoencoder to perform basic and detailed feature decomposition and source image reconstruction, achieving satisfactory fusion performance. CDDFuse effectively avoids the design of fusion rules by incorporating reversible network adaptive fusion features. In contrast, SDNet~\cite{Zhang2021SDNetAV} cascades a decomposition module on top of an encoder-decoder, decomposing the fused image closer to the source image. However, SDNet inevitably results in information loss. GeSeNet~\cite{Li2023GeSeNetAG} cascades a pretrained semantic module to complement the missing details of the fused image during the training phase.
In addition, GAN-based models are two-stage approaches. DDcGAN~\cite{Ma2020DDcGANAD} employs two discriminators to discriminate structural differences between a fused image and a source image pair. However, source image details may be lost during downsampling. Similarly, Zhou et al.~\cite{Zhou2022UnifiedGA} distinguished between gradient and intensity information through an input comprising an image processed by a gradient operator and an averaging filtering operation.
The discriminator in GS-MR-Fusion~\cite{Liu2022GliomaSM} is a segmentation network designed to determine whether a tumor region can be segmented from the fused image; this helps guide the model toward generating more meaningful results. However, these methods encounter difficulties in effectively addressing the instability of the training processes.

\subsection{Diffusion model}
\label{section2B}
Diffusion models~\cite{Ho2020DenoisingDP} are generative models that differ from previous GANs in that they simulate the diffusion process found in physics. They employ a distinctive workflow in which noise is gradually added to the data in a forward process and then progressively reduced by aligning the predicted noise with the actual added noise. This workflow allows diffusion models to generate high-quality, diverse samples, making them highly desirable for generative tasks.

As diffusion models mature, they have achieved notable success in many fields, including object detection~\cite{Wu2024DiffusionbasedNF} and image segmentation~\cite{Li2024CorrDiffCD}. They can generate highly realistic images and videos, including text-to-image~\cite{Czerkawski2023ExploringTC} and image editing~\cite{Yang2023EliminatingCP}. In addition, in superresolution reconstruction~\cite{Niu2023ACDMSRAC} tasks, diffusion models recover image details and improve image quality by fine-tuning the noise level. Nevertheless, diffusion models present obstacles, including slow processing speeds, challenges in pattern recognition from low-quality data, and an inability to generalize to new scenes~\cite{Cao2022ASO}.

Diffusion models have shown excellent performance in image fusion. DDFM~\cite{Zhao2023DDFMDD} represents a unified fusion approach.
It transforms the fusion problem into an excellent likelihood estimation problem with hidden variables. This process is then integrated into the diffusion process to generate conditional images.
FusionDiff~\cite{Li2023FusionDiffMI} is applied to multifocal image fusion, which employs ground truth–assisted training. In particular, the model introduces noise into the actual fused image during the forward process. The reverse process inputs the source image pairs and the noise-added ground truth. As the number of iterative steps in the reverse process increases, clear fused images are gradually generated. However, FusionDiff and DDFM involve a slow iterative sampling process, and the fusion efficiency is extremely low.
In contrast, Dif-Fusion~\cite{Yue2023DifFusionTH} uses only a diffusion process to train a noise prediction network, after which the network extracts features for fusion. This avoids the slow iterative process and significantly improves the fusion efficiency. However, Dif-Fusion is applied to infrared-visible image fusion, and it directly splices grayscale and color images to obtain diffusion features. Thus, it cannot perform fusion between grayscale images. The design of the feature fusion component in Dif-Fusion is relatively straightforward, making it difficult to leverage multiscale diffusion features fully.
On the basis of the above problems, we employed a diffusion process as a means of training feature extraction to improve the fusion efficiency. A unified fusion network is then proposed for application to various MIF tasks. Finally, fusion modules are constructed to adaptively integrate multimodal and multiscale diffusion features.

\section{Proposed Method}
\label{section3}
\subsection{Motivations}
MMIF is to combine the most meaningful information from disparate source images. MRI and CT image fusion mainly preserve the skull, calcification and tumor regions of CT images, soft tissues and other details of MRI images. MRI and functional image fusion preserve the details and edges of MRI images, as well as the high-brightness regions and color information of functional images. Because the two tasks' pixel intensities are inconsistent and existing networks have limited feature extraction capabilities, achieving uniform fusion with the same parameters is challenging.

This analysis aims to achieve unified fusion while balancing the brightness, color, contrast, and detail of the fused images. Specifically, UNet is trained to extract features by stabilizing a diffusion model in the inverse process. UNet forces the model to learn how to ignore noise while retaining useful information during progressive denoising, which improves its sensitivity and ability to capture critical details.
The multiscale feature of UNet allows the model to recover the detailed structure of an image at various levels. In addition, the fusion network incorporates three fusion modules designed for the adaptive and interactive integration of cross-modal and multiscale features tailored for different tasks.
In the loss function, the intensity of the pixels represents the overall luminance distribution, which reflects the contrast characteristics of the image. The difference between pixels constitutes the gradient, represented as the details in the image~\cite{Zhang2021SDNetAV}. Therefore, the hybrid loss is designed via a proportional combination strategy at the pixel, gradient, and structural levels to balance the luminance and detail information of the fused image.

We conducted numerous subjective and objective experiments with nine contrast methods on MRI-CT, MRI-PET, and MRI-SPECT datasets. We also performed additional experiments on phase-contrast (PC) and green fluorescent protein (GFP) image fusion without fine-tuning. Moreover, we extended it to infrared-visible image fusion (IVIF) by fine-tuning the fusion network.

\begin{figure*}[!t]
\centering
\includegraphics[width=6.3in]{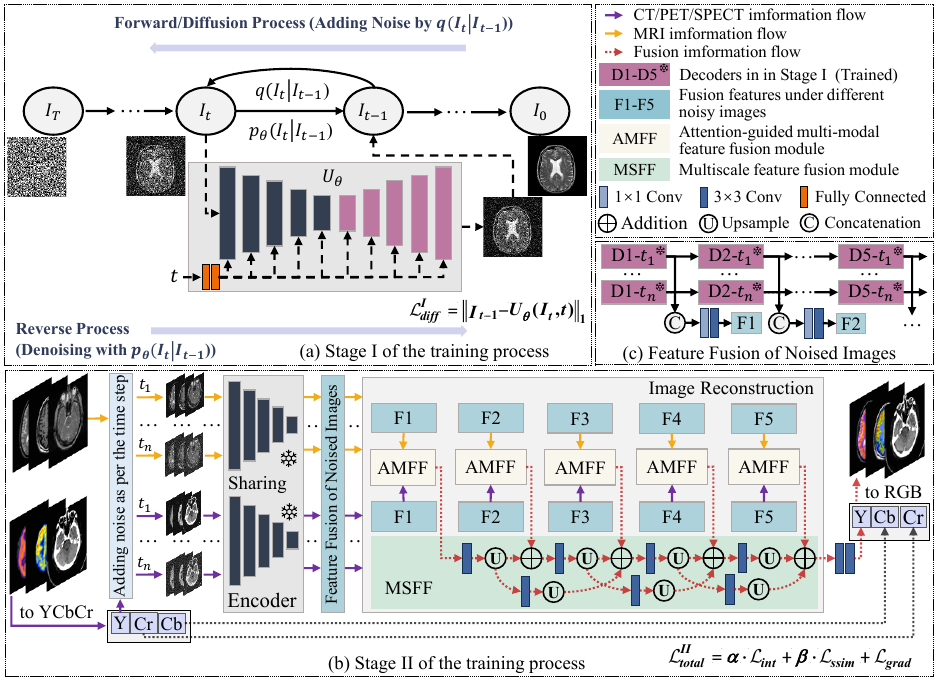}
\caption{Overall architecture of the diffusion model-based fusion network (DM-FNet). In Stage I, UNet is trained to reconstruct images through the diffusion process. In Stage II, the fusion network comprises three components: feature fusion under different noisy images, attention-guided multimodal feature fusion (AMFF), and multiscale feature fusion (MSFF).}
\label{Fig2}
\vspace{-0.4cm}
\end{figure*}

\subsection{Overall architecture}
The proposed DM-FNet is a two-stage model comprising a diffusion model-based reconstruction framework and a fusion network. The overall framework is shown in Fig. \ref{Fig2}. Data preprocessing requires color space transformation of PET and SPECT images to achieve unified fusion and recover superior color information, facilitating extended fusion of single-channel images.
Specifically, the diffusion model is applied to MRI, CT, PET, and SPECT images as inputs in the initial stage (Fig. \ref{Fig2}(a)). In the forward process, Gaussian noise is gradually added to the data until it approaches pure noise. The reverse process captures the detailed features of the image by progressively reducing the noise during the reconstruction of UNet, providing a rich feature representation for the fusion network. This stage ensures accurate restoration of details in medical images, particularly for critical structures such as tumors and blood vessels.
In Stage II, the extracted diffusion features under different noise-added images are input into the fusion network (Fig. \ref{Fig2}(b)). The fusion network considers three aspects: multiple inputs, multiple modalities, and multiple scales. This design enables the network to adapt to different datasets without specific fusion rules. Adaptive fusion features ensure that the fused image retains the strengths of each modality, such as the soft tissue structures from MRI, calcification regions from CT, and radioactive tracer distribution from PET.
Finally, a hybrid loss function is introduced, incorporating the intensity, gradient, and structural levels. This loss function guides the network to balance the recovery of details (e.g., tumor boundaries, brain sulci) and preserve image brightness consistency, ultimately providing clinicians with clearer and more diagnostically useful images.

\subsection{Stage I: diffusion process-based reconstruction framework}
\label{section3A}
It is assumed that the input to the reconstruction framework is $I_0\ \in\{{I_{MRI},I}_{CT},I_{PET},I_{SPECT}\}$. A denoising diffusion probabilistic model is employed to learn the underlying structures of medical images. The forward process gradually adds noise at $T$ time steps, and the reverse process progressively removes noise by UNet, which predicts images from the previous time step.

{{\bf{Forward process.}} Gaussian noise with $T$ time steps is gradually incorporated into the dataset until it reaches a state of pure noise \cite{Ho2020DenoisingDP}. The noise image $I_t$ at time step $t$ can be calculated from the previous time step image $I_{t-1}$ as follows:
\begin{equation}
\label{eq1}
q_{t}(I_{t}|I_{t-1})=\mathcal{N}(I_{t};\sqrt{\alpha_{t}}I_{t-1},(1-\alpha_{t}))Z
\end{equation}
where $Z$ denotes the standard normal distribution. $I_t$ and $I_{t-1}$ denote the noise images generated at time steps $t$ and $t-1$, respectively ($t=1,2,...,T$). $\alpha_t$ is the variance scheduling used to control Gaussian noise. Given the sampled noise and variance scheduling $\alpha_1$, ..., $\alpha_T$, the noise samples at time step $t$ can be computed directly from the input image $I_0$.
\begin{equation}
\label{eq2}
q_{t}(I_{t}|I_{0})=\mathcal{N}(I_{t};\sqrt{\prod_{i}^{t}\alpha_{i}}\cdot
I_{0},(1-\prod_{i}^{t}\alpha_{i}))Z
\end{equation}

{\bf{Reverse process.}} To recover meaningful structures from noisy data, the denoising reconstruction network $U_\theta(,)$ directly predicts the image of the previous time step rather than the added noise. Specifically, the structure of $U_\theta(,)$ adopts the UNet structure reported in the literature~\cite{Saharia2021ImageSV}, which contains five scales of diffusion features.
\begin{equation}
\label{eq3}
I_{t-1}^{\prime}=U_{\theta}(I_t,t)
\end{equation}

\subsection{Stage II: fusion network}
\bf{Feature fusion of different noisy images.}} In the diffusion inverse process, the inputs at each step are noisy images with varying noise levels. Therefore, the model must be fine-tuned to capture additional details. To this end, we use Stage I trained $U_\theta(,)$ to extract noise features at different time steps as the input to the fusion network. Given the original images $I_A=I_{MRI}$ and $I_B\in\{I_{CT},I_{PET},I_{SPECT}\}$, the images after adding time step $t$ are $I_{A,t}$ and $I_{B,t}$, where $t\in\{t_1,...,t_n\}$ and $t\in[0,T]$. The feature map after the encoder $E$ is calculated as follows:
\begin{equation}
\label{eq04}
H_{A,t}^{D_0}=E\left(I_{A,t}\right),\ H_{B,t}^{D_0}=E\left(I_{B,t}^Y\right)
\end{equation}
\begin{equation}
\label{eq05}
H_{A,t}^{D_i}=\mathcal{D}_i\left(H_{A,t}^{D_{i-1}}\right),H_{B,t}^{D_i}=\mathcal{D}_i\left(H_{B,t}^{D_{i-1}}\right)
\end{equation}
where $i=1,.... ,5$. The initial fusion of the extracted features is performed, and the formula is calculated as follows:
\begin{equation}
\label{eq06-1}
f_{A,t}^i=\varphi_3(\varphi_1([H_{A,t_1}^{D_i},...,H_{A,t_n}^{D_i}]))
\end{equation}
\begin{equation}
\label{eq06-2}
f_{B,t}^i=\varphi_3(\varphi_1([H_{B,t_1}^{D_i},...,H_{B,t_n}^{D_i}]))
\end{equation}
where $\varphi_3$ is a 3 × 3 conv and where $\varphi_1$ is a 1 × 1 conv. After the above steps, we obtain feature maps at different scales in different modes.

{\bf{Attention-guided multimodal feature fusion (AMFF).}} We introduce an AMFF module~\cite{Chen2023DEANetSI}, designated $\mathcal{A}$, designed to extract and fuse key information from different modalities in a hierarchical and fine-grained manner. The module obtains the weight maps of the input feature maps in a coarse-to-fine manner, ensuring sufficient blending of the channel and spatial attention to facilitate information interaction. The AMFF is defined as follows:

\begin{equation}
\label{eq07}
F^i=\mathcal{A}\left(f_{A,t}^i,f_{B,t}^i\right)
\end{equation}

\begin{figure}[!t]\centering
\includegraphics[width=2.7in]{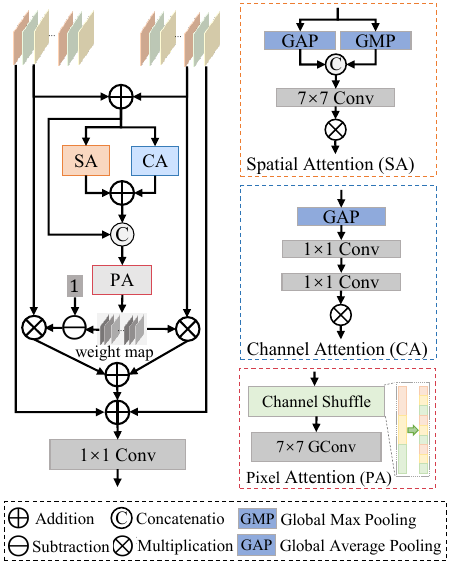}
\caption{Network structure of the AMFF.}
\label{Fig4}
\vspace{-0.4cm}
\end{figure}

As illustrated in Fig. \ref{Fig4}, spatial attention (SA) and channel attention (CA) are applied independently to the input feature map to identify and reinforce the most critical local regions and feature channels.
Specifically, SA effectively focuses on lesion areas, tumors, or key anatomical structures. Moreover, CA helps the model emphasize important features from different modalities (e.g., brain tissue in MRI, metabolically active regions in PET images).
Although these two attention mechanisms have distinct foci, the information they reveal is complementary.
Consequently, we integrate them via a weighted summation operation to assign a weight to each feature element that combines spatial and channel importance.

To prevent potential information loss, we maintain the integrity of the original data via feature map splicing. A pixel attention (PA) mechanism is then introduced. This mechanism generates a refined weight map for each feature map, thereby guiding the model to focus more precisely on key pixel points in the image, especially at the pixel-level details of lesion areas, tumor boundaries, or important organ regions.
In this process, channel shuffling facilitates the exchange of information between different feature channels, whereas group convolution decreases the use of computational resources without compromising performance.

Ultimately, a fused feature map that contains rich details and highlights key information is obtained by combining the original feature map with its corresponding weight map.

{\bf{Multiscale feature fusion (MSFF)}.} The multiscale nature of UNet allows the model to recover the detailed structure of the image at different levels. To this end, we utilize the MSFF module to integrate feature maps across five scales. Given that the numbers of feature maps and scales differ, mapping is performed via 3 × 3 conv and upsampling. The specific operation is as follows:
\begin{equation}
\label{eq7}
M^1=U_p(\varphi_3(F^1))
\end{equation}
\begin{equation}
\label{eq8}
M^2=U_p(\varphi_3(M^1+F^2))
\end{equation}
\begin{equation}
\label{eq9}
M^i=U_p(\varphi_3(M^{i-1}+F^{i-1}+U_p(\varphi_3(M^{i-2})))
\end{equation}
where $i=3,4,5$ and where $U_p$ denotes upsampling. Finally, the final fused image $I_F$ is obtained via Eq. (\ref{eq10}).
\begin{equation}
\label{eq10}
I_F=\varphi_3(\varphi_3(M^5))
\end{equation}

\subsection{Loss fuction}
\label{section3C}
{\bf{Stage I}.} The purpose of $U$ in the diffusion model is to remove the noise of a single step added in $I_t$, thereby enabling the prediction of $I_{t-1}$. Thus, the loss function for Stage I can be formulated as follows:
\begin{equation}
\label{eq15}
\mathcal{L}_{diff}^I=\frac{1}{HW}\cdot\|I_{t-1}-U_\theta(I_t,t)\|_1
\end{equation}
where $\|\cdot\|_1$ denotes the $L_1$ norm. $H$ and $W$ denote the height and width of the image, respectively.

{\bf{Stage II}.} The loss function of the fusion network combines intensity, structural, and gradient levels, defined as follows:
\begin{equation}
\label{eq01}
{\mathcal L}_{total}^{II}=\alpha\cdot{\mathcal L}_{int} + \beta\cdot {\mathcal L}_{ssim} + {\mathcal L}_{grad}
\end{equation}
where $\alpha$ and $\beta$ are the weights of the loss functions, which are discussed in detail in Section \ref{section4B}.

We want the fused image to focus more on the high-brightness region. Thus, we select the highest pixel value from the source image pair to compute the loss value with the fused image. The intensity loss is defined as follows:
\begin{equation}
\label{eq02}
{\mathcal L}_{int}=\frac{1}{HW}\cdot\|I_{F}^Y-max(I_{A}, I_{B}^Y)\|_1
\end{equation}

This paper introduces a structure similarity (SSIM) loss function driven by the maximum local standard deviation (std) \cite{Ding2023M4FNetMM}. This loss function calculates the final loss value by comparing the local standard deviation of selected regions in the source image. This helps prevent the model from being biased toward images with fewer effective regions and reduces the need for additional parameters. The formula for ${\mathcal L}_{ssim\_std}$ is shown in Eqs. (\ref{eq17}) and (\ref{eq19}).
\begin{equation}
\label{eq17}{\mathcal L}_{ssim}=1-\frac1N\sum_{P=1}^NSSIM(I_{F,P}^Y,S(I_{A,P}^Y,I_{B,P}^Y))
\end{equation}
\begin{equation}
\label{eq19}S(I_{A,P}^Y,I_{B,P}^Y)=\begin{cases}I_{A,P}^Y,if std(I_{A,P}^Y)>std(I_{B,P}^Y)\\I_{B,P}^Y,if std(I_{A,P}^Y)<std(I_{B,P}^Y)\end{cases}
\end{equation}
where $I_{F,P}^Y$ represents the $P$-th patch of the luminance channel of the fused image, with sizes of $w{\times}w$, and where $I_{A,P}^Y$ and $I_{B,P}^Y$ are similar to each other. $SSIM(\cdot)$ calculates the structural similarity of corresponding positions between the specified regions of the image. Furthermore, the selection of the loss region on the basis of the local standard deviation of the $P$-th patch is denoted as $S(I_{A,P}^Y,I_{B,P}^Y)$.

In gradient loss, the texture details of the fused image should align with the most prominent texture details observed in the corresponding region of the source image. The calculation formula is as follows:
\begin{equation}
\label{eq03}
\mathcal{L}_{\text {grad }}=\frac{1}{H W} \cdot\left(\left\|\left|\nabla I_{F}^Y\right|-\max \left(\left|\nabla I_{A}\right|,\left|\nabla I_{B}^{Y}\right|\right)\right\|_{1}\right)
\end{equation}
where $\mathrm{\nabla}$ denotes the Sobel gradient operator.

\section{Experiments}
\label{section4}
\subsection{Preliminary work}
\label{section4A}
{\bf{Datasets.}} This paper utilizes images from a publicly available dataset provided by Harvard Medical School (\href{https://www.med.harvard.edu/AANLIB/home.html}{https://www.med.harvard.edu/AANLIB/home.html}). The dataset includes various modalities of brain disease images, each measuring $256\times256$ pixels. This study employed four imaging modalities: MRI, CT, PET, and SPECT. The training set consisted of 30 pairs each of MRI-CT, MRI-PET, and MRI-SPECT images, totaling 90 pairs. The three test sets each contained 50 pairs of images for MRI-CT, MRI-PET, and MRI-SPECT. The MRI included T1-weighted imaging (MR-T1), T2-weighted imaging (MR-T2), and MR-Gad (MR-T1 after enhancement with the contrast media Gd-DTPA).

{\bf{Evaluating indicators.}} To assess the quality of fused images, nine objective indicators were selected based on the characteristics of MMIF: spatial frequency (SF)~\cite{Eskicioglu1995ImageQM}, standard deviation (SD)~\cite{Rao1997InfibreBG}, average gradient (AG)~\cite{Cui2015DetailPF}, Piellas metric $Q_W$~\cite{Piella2003ANQ}, sum of the correlations of differences (SCD)~\cite{Aslanta2015ANI}, visual information fidelity for fusion (VIFF)~\cite{Han2013ANI}, normalized weighted edge information $Q_{AB/F}$~\cite{Xydeas2000ObjectiveIF}, multiscale structural similarity (MSSSIM)~\cite{Wang2003MultiscaleSS}, and feature mutual information with wavelet transform (FMI-WT)~\cite{Haghighat2011ANI}.

{\bf{Comparative methods.}} We conducted comparative experiments using two traditional methods: TL-SR (2021)~\cite{Li2021JointIF} and Cloud (2022)~\cite{Wang2022MultimodalMI}),  and seven state-of-the-art (SOAT) deep learning methods: (DSAGAN (2020)~\cite{Fu2021DSAGANAG}, EMFusion (2021)~\cite{Xu2021EMFusionAU}, SDNet (2021)~\cite{Zhang2021SDNetAV}, MATR (2022)~\cite{Tang2022MATRMM}, CDDFuse (2022)~\cite{Zhao2022CDDFuseCD}, MsgFusion (2023)~\cite{Wen2024MsgFusionMS}, and LRFNet (2024)~\cite{He2024LRFNetAR}).

{\bf{Implementation details.}} DM-FNet was implemented via the PyTorch framework, and all training was performed on four NVIDIA GeForce RTX 3090 GPUs. All methods were tested on a 3.60 GHz Intel® Core™ i9-9900K CPU, 64 GB of RAM, and a single NVIDIA GeForce RTX 2080 Ti GPU. Furthermore, the time steps described in Section \ref{section3A} are defined as $t_1=5$, $t_2=10$, and $t_3=20$. The hyperparameters of the loss function are set to $\alpha=1.5$ and $\beta=0.5$.

\subsection{Ablation Experiments} 
\label{section4B}
\begin{figure*}[ht]\centering
\includegraphics[width=7.0in]{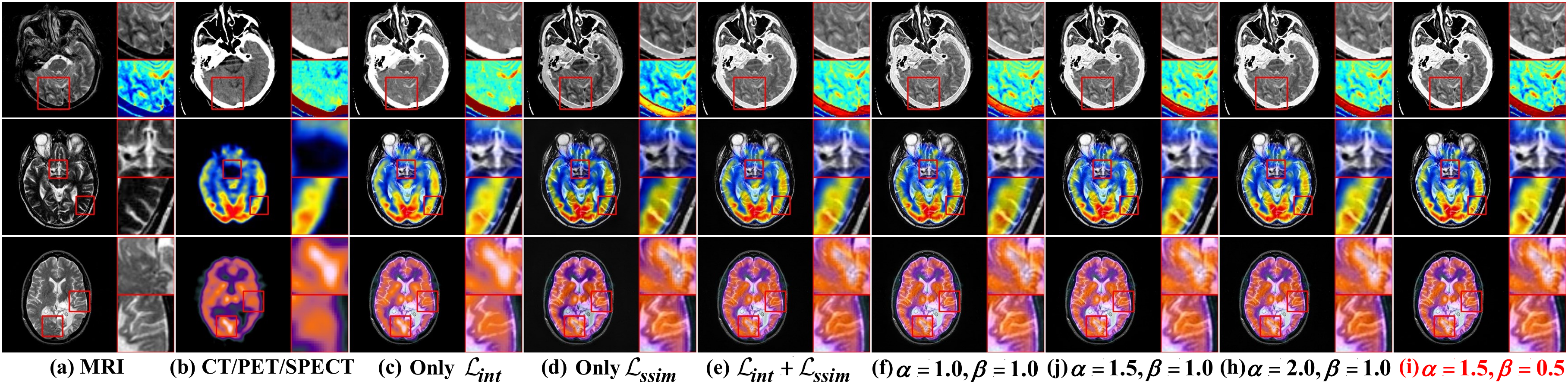}
\caption{Fusion results for different values of $\alpha$ and $\beta$.}
\label{Fig_LOSS}
\end{figure*}

\begin{table*}[!ht]
    \centering
    \renewcommand\arraystretch{1}
    \caption{Objective evaluation for different values of $\alpha$ and $\beta$ (averaged over three test sets). Red: best, blue: second best.}
    \label{tb_loss}
    \begin{tabular}{@{}llllllllll@{}}\hline
		Experiments & SF & SD & AG & $Q_W$ & SCD & VIFF & $Q_{AB/F}$ & MSSSIM & FMI\_WT \\ \hline
        Only ${\mathcal L}_{int}$ & 27.933  & \textcolor{red}{77.543}  & 7.165  & 0.805  & 1.445  & 0.510  & 0.609  & 0.906  & \textcolor{red}{0.313} \\
		Only ${\mathcal L}_{ssim}$ & {28.428}  & 64.858  & {7.939}  & {0.850}  & 1.017  & 0.529  & \textcolor{red}{0.670}  & \textcolor{red}{0.950}  & 0.251 \\
		${\mathcal L}_{int}+{\mathcal L}_{ssim}$ & 27.683  & 72.881  & 7.591  & 0.844  & 1.372  & 0.513  & 0.644  & 0.940  & 0.241 \\
		$\alpha=1.0, \beta=1.0$ & \textcolor{blue}{31.464}  & 74.472  & \textcolor{red}{8.728}  & 0.850  & 1.443  & 0.527  & \textcolor{blue}{0.661}  & 0.937  & 0.270 \\
	    $\alpha=1.5, \beta=1.0$ & 30.254  & 76.514  & 8.333  & \textcolor{blue}{0.851}  & \textcolor{red}{1.521}  & \textcolor{blue}{0.545}  & 0.650  & \textcolor{blue}{0.941}  & {0.293} \\ 
		$\alpha=2.0, \beta=1.0$ & 29.377  & 75.630  & 8.172  & 0.851  & {1.460}  & 0.529  & {0.658}  & 0.936  & 0.269 \\
	    $\alpha=1.5, \beta=0.5$ & \textcolor{red}{31.577}  & \textcolor{blue}{77.328}  & \textcolor{blue}{8.489}  & \textcolor{red}{0.851}  & \textcolor{blue}{1.485}  & \textcolor{red}{0.545}  & 0.649  & 0.936  & \textcolor{blue}{0.312} \\ \hline
    \end{tabular}
\vspace{-0.2cm}
\end{table*}
\begin{table*}[ht]
\renewcommand\arraystretch{1}
\begin{center}
\caption{Objective evaluation for different time-step combinations (averaged over three test sets). Red: best, blue: second best.}
\label{tbl3}
    \begin{tabular}{@{}llllllllll@{}}\hline
        Time steps & SF & SD & AG & $Q_W$ & SCD & VIFF & $Q_{AB/F}$ & MSSSIM & FMI\_WT  \\ \hline
        (5) & 30.700  & 75.691  & 8.477  & 0.839  & 1.446  & {0.530}  & 0.637  & 0.931  & 0.270   \\
        (10) & 30.600  & 75.684  & 8.410  & 0.844  & 1.439  & \textcolor{blue}{0.533}  & 0.633  & 0.936  & 0.266   \\
        (20) & 30.482  & \textcolor{blue}{76.71}2  & 8.469  & 0.826  & \textcolor{blue}{1.449}  & 0.523  & 0.573  & 0.921  & 0.246   \\
        (50) & 29.781  & 75.982  & 8.328  & 0.831  & 1.442  & 0.522  & 0.579  & 0.925  & 0.242   \\
        (5, 10) & \textcolor{blue}{31.330}  & 75.508  & \textcolor{red}{8.499}  & \textcolor{blue}{0.855}  & 1.435  & 0.522  & \textcolor{red}{0.667}  & 0.933  & \textcolor{blue}{0.289}   \\
        (5,10,20) & \textcolor{red}{31.577}  & \textcolor{red}{77.328}  & \textcolor{blue}{8.489}  & {0.851}  & \textcolor{red}{1.485}  & \textcolor{red}{0.545}  & {0.649}  & \textcolor{red}{0.936}  & \textcolor{red}{0.312}   \\
        (5,10,20,50) & 29.386  & {73.682}  & 8.040  & \textcolor{red}{0.855}  & 1.387  & 0.513  & \textcolor{blue}{0.662}  & \textcolor{blue}{0.936}  & 0.269   \\ \hline
    \end{tabular}
\end{center}
\vspace{-0.5cm}
\end{table*}

{\bf{Hyperparametric analysis.}}
This section evaluates the effects of different hyperparameter values in the loss function on the fusion performance from subjective and objective perspectives. Fig. \ref{Fig_LOSS} and Table \ref{tb_loss} reveal several important conclusions:
1) Under the constraint of $\mathcal{L}_{int}$, the model perfectly preserves the high-pixel areas of the CT and functional images but loses details in MRI images, as illustrated in Fig. \ref{Fig_LOSS}(c).
2) Under the constraint of $\mathcal{L}_{ssim}$, the recovery of high-pixel areas is poor, as shown in Fig. \ref{Fig_LOSS}(d).
3) Combining $\mathcal{L}_{int}$ with $\mathcal{L}_{ssim}$, Fig. \ref{Fig_LOSS}(e) shows a balanced preservation of high-pixel regions and details. However, restoring areas such as the skull in the CT image remains suboptimal.
4) After adding $\mathcal{L}_{grad}$, Table \ref{tb_loss} shows a notable improvement in the detail of the AG metric. The visual enhancement of details is also apparent, as illustrated in Fig. \ref{Fig_LOSS}(f).

Based on Fig. \ref{Fig_LOSS}(e) and \ref{Fig_LOSS}(f), when the weights of $\mathcal{L}_{int}$ and $\mathcal{L}_{ssim}$/$\mathcal{L}_{grad}$ are equal, the restoration of high-pixel areas in CT images is suboptimal. Consequently, we further explored the impact of increasing the effects of either increasing the weight of $\mathcal{L}_{int}$ or decreasing the weight of $\mathcal{L}_{ssim}$ on the fusion performance. Fig. \ref{Fig_LOSS}(g)-(i) demonstrate that with these three parameter settings, the fusion images effectively preserve both high-pixel areas and detailed information. After considering the visual effects and quantitative analysis, we choose $\alpha=1.5$ and $\beta=0.5$.

{\bf{Different time-step combinations.}}
The proposed DM-FNet integrates the diffusion features at various time intervals. Therefore, we examined the principles of combining data from one to multiple time steps using only the seven examples presented in this section.
The results in Table \ref{tbl3} lead us to two conclusions. First, the model's performance declines as time progresses within a single time step. Second, the model performs better with increasing time steps, with the combination of (5, 10, 20) yielding the best results.
In summary, carefully combining multiple diffusion features at different time steps improved model performance to some extent. However, the computational resource overhead also increased.

\begin{figure}[!t]\centering
\includegraphics[width=2.3in]{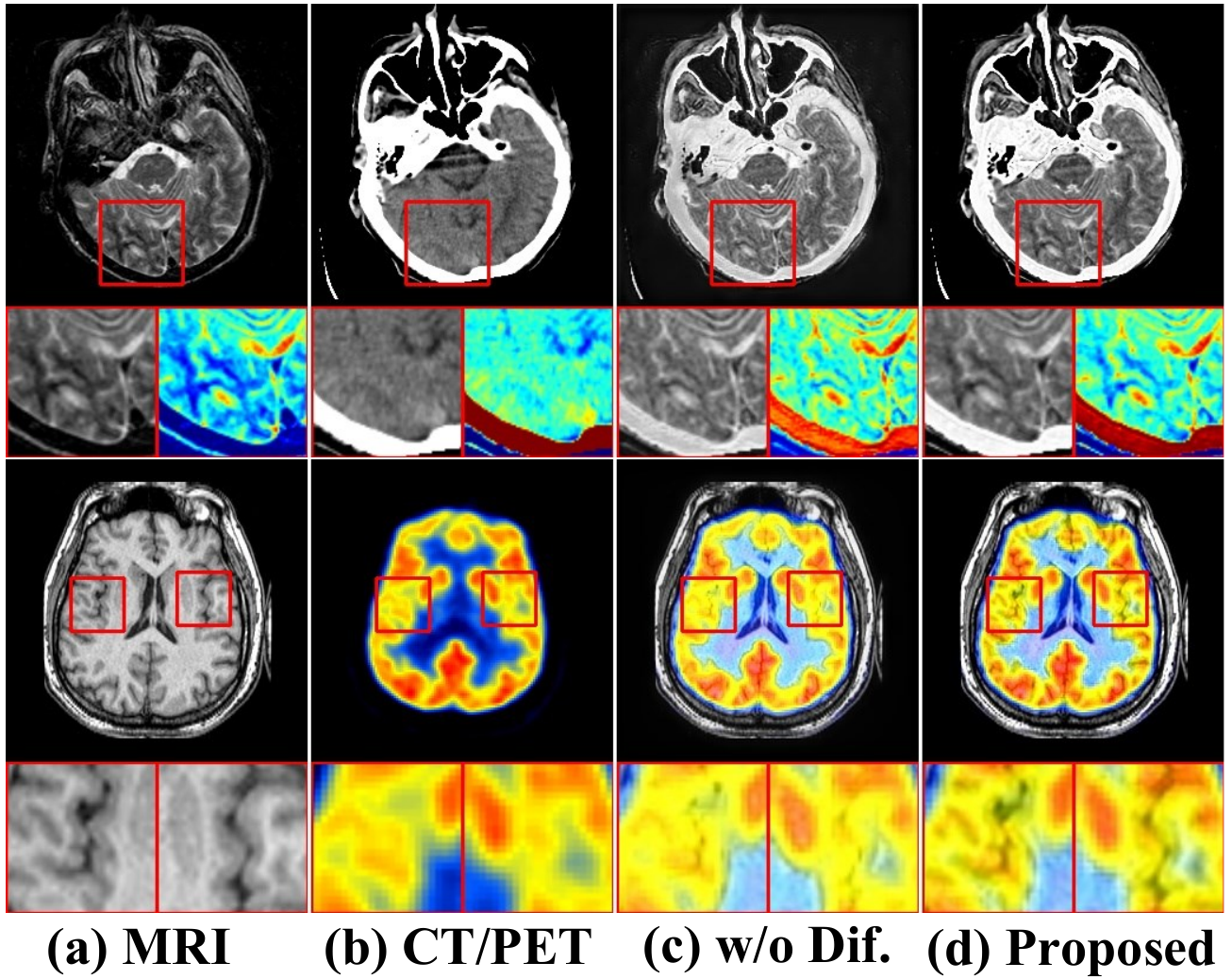}
\caption{Visualization analysis without diffusion process.}
\label{Fig5}
\vspace{-0.5cm}
\end{figure}

{\bf{Analysis of the diffusion process.}} To verify the validity of the diffusion model, this section eliminates the diffusion process. In Stage I, the same UNet architecture and loss function were used to establish an image reconstruction framework. In Stage II, the fusion module of different noise conditions was removed, whereas the other modules remained unchanged. The results shown in Fig. \ref{Fig5} indicate that removing the diffusion process significantly degrades the model's ability to retain details. Most of the evaluation metrics across all test sets in Table \ref{tbl4-1} were also noticeably reduced, demonstrating that the absence of the diffusion process substantially impacted feature extraction capabilities.

\begin{table*}[!t]
\renewcommand\arraystretch{1}
\begin{center}
\caption{Objective evaluation results without diffusion process. Red: best results.}
\label{tbl4-1}
    \begin{tabular}{@{}llllllllllll@{}}\hline
         Modality & Experiments & SF & SD & AG & $Q_W$ & SCD & VIFF & $Q_{AB/F}$ & MSSSIM & FMI\_WT  \\ \hline
        MRI-CT & w/o Dif. & {32.188}  & {88.067}  & {8.367}  & {0.683}  & \textcolor{red}{1.479}  & {0.412}  & {0.524}  & {0.892}  & {0.249}  \\
        ~ & Proposed & \textcolor{red}{38.759}  & \textcolor{red}{89.708}  & \textcolor{red}{9.000}  & \textcolor{red}{0.744}  & {1.440}  & \textcolor{red}{0.436}  & \textcolor{red}{0.571}  & \textcolor{red}{0.920}  & \textcolor{red}{0.379}  \\
        MRI-PET & w/o Dif. & {25.697}  & {72.426}  & {7.809}  & {0.851}  & \textcolor{red}{1.571}  & {0.542}  & {0.626}  & {0.920}  & {0.230}  \\ 
        ~ & Proposed & \textcolor{red}{29.258}  & \textcolor{red}{73.751}  & \textcolor{red}{8.699}  & \textcolor{red}{0.894}  & {1.570}  & \textcolor{red}{0.581}  & \textcolor{red}{0.668}  & \textcolor{red}{0.940}  & \textcolor{red}{0.269}  \\ 
        MRI-SPECT & w/o Dif. & {25.490}  & {67.306}  & {7.760}  & {0.885}  & \textcolor{red}{1.495}  & {0.580}  & {0.671}  & {0.921}  & {0.236}  \\ 
        ~ & Proposed & \textcolor{red}{26.713}  & \textcolor{red}{68.524}  & \textcolor{red}{7.769}  & \textcolor{red}{0.917}  & {1.446}  & \textcolor{red}{0.619}  & \textcolor{red}{0.708}  & \textcolor{red}{0.950}  & \textcolor{red}{0.287}  \\ \hline
    \end{tabular}
\end{center}
\vspace{-0.2cm}
\end{table*}

{\bf{Analysis on different modules.}} 
This section verifies the validity of the MSFF and AMFF, which utilize nine evaluation metrics, along with the model's parameters and floating point operations (FLOPs). Specifically, "w/o AMFF" refers to the fusion of multimodal features via the addition operation, whereas "w/o MSFF" involves the use of only the features from the final layer.
Table \ref{tbl4} shows that removing any module results in a significant performance drop. SF, SD, AG, $Q_W$, SCD, and MS-SSIM directly reflect image sharpness, detail preservation, and structural integrity. Among them, AMFF and MSFF emphasize cross-modal feature complementarity and multiscale structural consistency, potentially suppressing nonsalient edges and causing slight degradation in $Q_{abf}$, particularly in edge-aware computing. Brain CT images are characterized by high-contrast bony structures (e.g., the skull), where poor reconstruction in these regions can lead to elevated AG values.
Moreover, the model's parameters and FLOPs for Stage I are 20,648 KB and 194.816 G, respectively. As shown in Table \ref{tbl4}, the parameters and FLOPs of the fusion network are significantly lower than those of Stage I.
\begin{table*}[!t]
\renewcommand\arraystretch{1}
\begin{center}
\caption{Objective evaluation results without MSFF and AMFF. Red: best results.}
\label{tbl4}
    \begin{tabular}{@{}llllllllllllll@{}}\hline
         Modality & Experiments & SF & SD & AG & $Q_W$ & SCD & VIFF & $Q_{AB/F}$ & MSSSIM & FMI\_WT & Param(KB) & FLOPs(G)  \\ \hline
        MRI-CT & w/o MSFF. & {38.652}  & {88.487}  & \textcolor{red}{9.237}  & {0.744}  & {1.430}  & {0.427}  & {0.586}  & {0.917}  & {0.359}  & \textcolor{red}{245} & \textcolor{red}{15.773} \\
        ~ & w/o AMFF. & {37.577}  & {86.444}  & {9.099}  & {0.742}  & {1.415}  & {0.418}  & \textcolor{red}{0.588}  & {0.914}  & {0.293}  & {2978} & {57.328}  \\
        ~ & Proposed & \textcolor{red}{38.759}  & \textcolor{red}{89.708}  & {9.000}  & \textcolor{red}{0.744}  & \textcolor{red}{1.440}  & \textcolor{red}{0.436}  & {0.571}  & \textcolor{red}{0.920}  & \textcolor{red}{0.379}  & {3027} & {57.549}   \\
        MRI-PET & w/o MSFF. & {28.139}  & {72.468}  & {8.560}  & {0.892}  & {1.555}  & {0.564}  & \textcolor{red}{0.669}  & {0.939}  & \textcolor{red}{0.274}  & \textcolor{red}{245} & \textcolor{red}{15.773}  \\ 
        ~ & w/o AMFF. & {27.540}  & {70.500}  & {8.438}  & {0.891}  & {1.507}  & {0.549}  & {0.661}  & {0.938}  & {0.226}  & {2978} & {57.328}  \\
        ~ & Proposed & \textcolor{red}{29.258}  & \textcolor{red}{73.751}  & \textcolor{red}{8.699}  & \textcolor{red}{0.894}  & \textcolor{red}{1.570}  & \textcolor{red}{0.581}  & {0.668}  & \textcolor{red}{0.940}  & {0.269}  & {3027} & {57.549}  \\ 
        MRI-SPECT & w/o MSFF. & {25.892}  & {67.887}  & {7.736}  & {0.910}  & {1.400}  & {0.606}  & \textcolor{red}{0.711}  & {0.947}  & {0.264}  & \textcolor{red}{245} & \textcolor{red}{15.773}   \\
        ~ & w/o AMFF. & {25.280}  & {66.251}  & {7.599}  & {0.910}  & {1.328}  & {0.588}  & {0.704}  & {0.947}  & {0.230}  & {2978} & {57.328}  \\
        ~ & Proposed & \textcolor{red}{26.713}  & \textcolor{red}{68.524}  & \textcolor{red}{7.769}  & \textcolor{red}{0.917}  & \textcolor{red}{1.446}  & \textcolor{red}{0.619}  & {0.708}  & \textcolor{red}{0.950}  & \textcolor{red}{0.287}  & {3027} & {57.549}  \\ \hline
    \end{tabular}
\end{center}
\vspace{-0.5cm}
\end{table*}

\subsection{Analysis of fusion results} 
\label{section4D}
\begin{figure*}[!htbp]
	\centering
	\begin{minipage}{1\linewidth}
		\centering
		\includegraphics[width=7.0in]{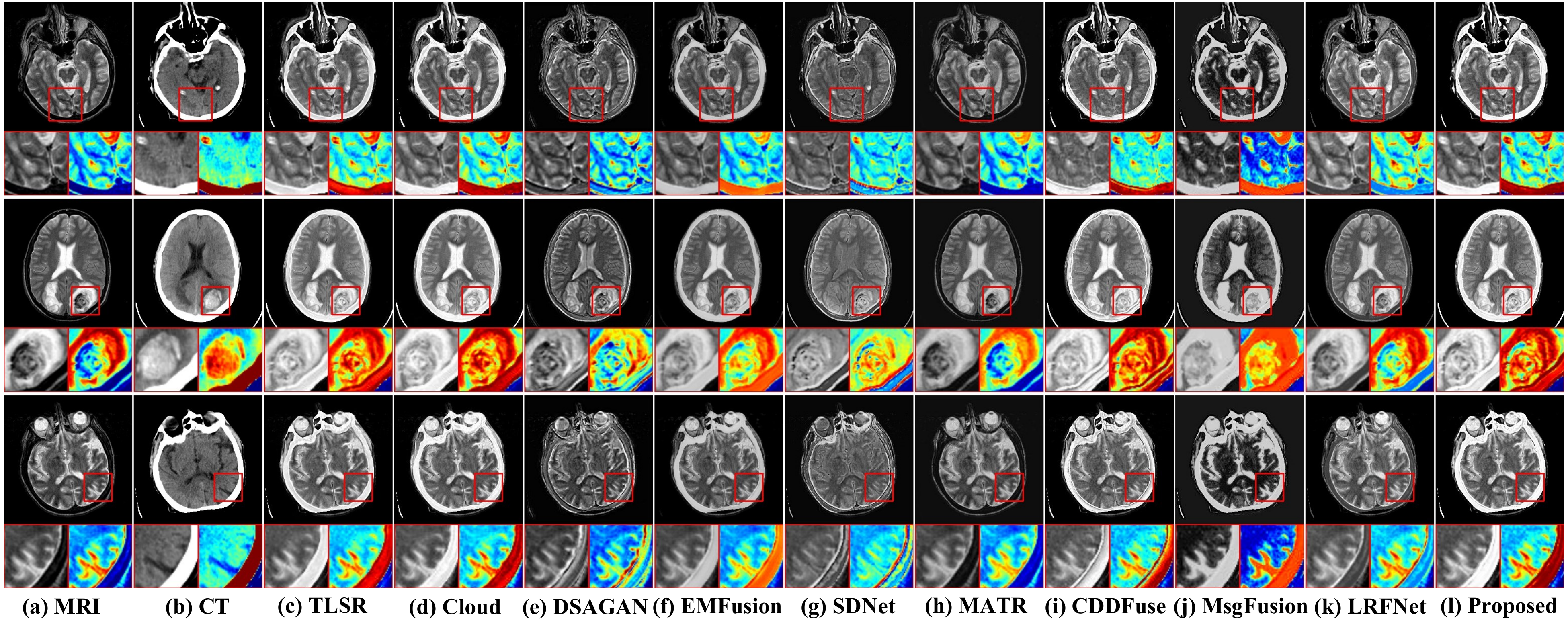}
		\caption{Visual comparison of DM-FNet with 9 SOAT methods for MRI and CT image fusion. The regions are enlarged for a more intuitive comparison, using close-ups and pseudocolor maps for better observation.}
		\label{Fig6}
	\end{minipage}
	\begin{minipage}{1\linewidth}
		\centering
		\includegraphics[width=7.0in]{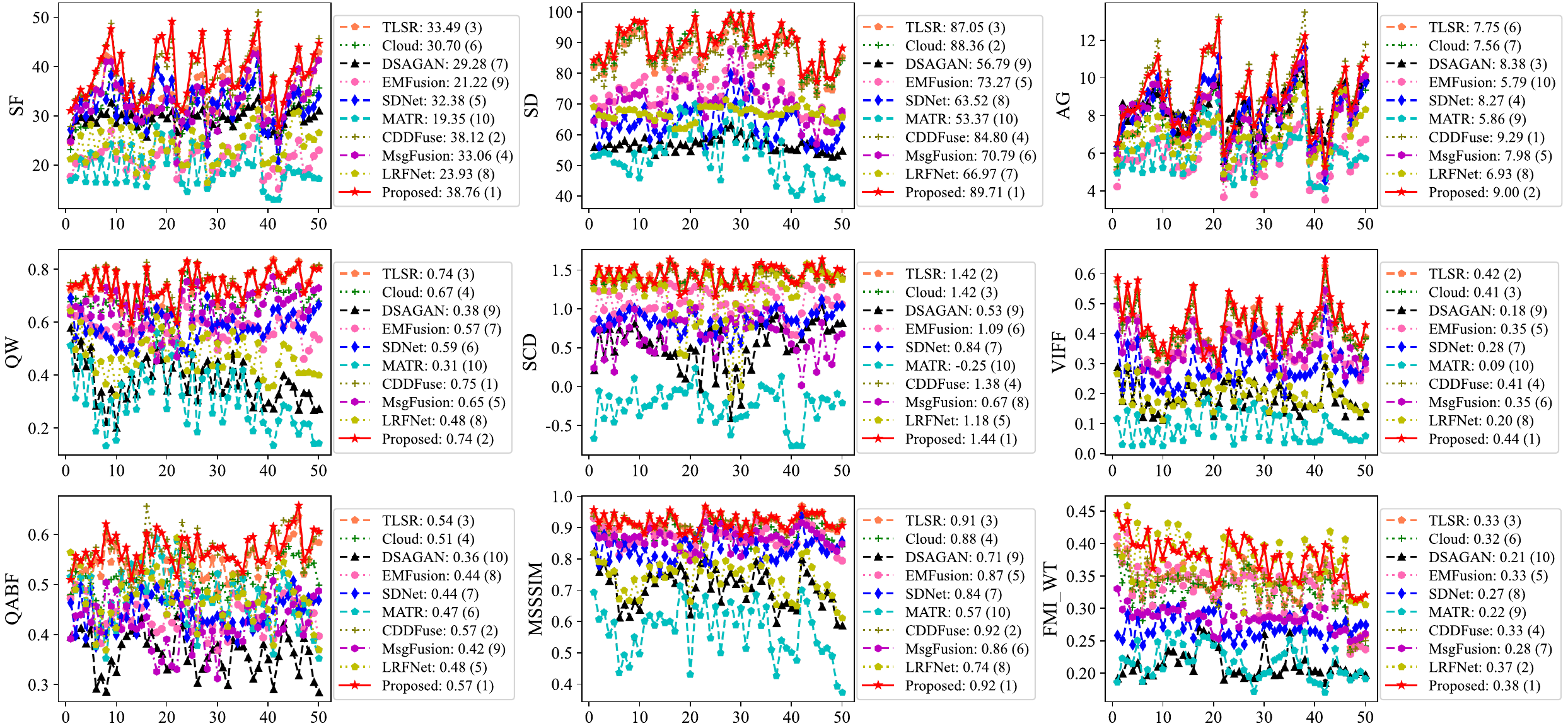}
		\caption{Quantitative comparison of DM-FNet with 9 SOAT methods for MRI and CT image fusion. The legend on the right displays the average scores for different methods.}
		\label{Fig7}
	\end{minipage}
\vspace{-0.3cm}
\end{figure*}

{\bf{MRI and CT image fusion.}}
As shown in Fig. \ref{Fig6}, we compared the fusion results of DM-FNet to those obtained by the nine SOAT methods on three sets of typical MRI and CT images.
DSAGAN, MATR, and LRFNet, which consider only MRI and functional image fusion, yield suboptimal fusion results for MRI–CT image fusion, particularly in the calcification and cranial regions.
SDNet's lightweight architecture resulted in a lack of robust feature representation, which is evident from its inability to clearly delineate the cranial region.
The fusion results of MsgFusion and EMFusion were significantly darker, with EMFusion exhibiting more black regions absent in the original images.
Compared with DM-FNet, CDDFuse was less sensitive to detailed information, resulting in slightly poorer texture and edge feature recovery in the fused images.
TLSR, Cloud, and DM-FNet produce fused images with rich details, effectively preserving calcification and the cranial region.

The results of the nine objective evaluation metrics on 50 image pairs are presented in Fig. \ref{Fig7}. The proposed DM-FNet achieved optimal values for SF, SD, SCD, VIFF, $Q_{AB/F}$, MSSSIM, and FMI-WT. Among these, SD and SCD were optimal, indicating that the fusion results of DM-FNet were more stable at the pixel level. Moreover, the fusion images displayed complete calcification and cranial areas, leading to slightly lower AG and $Q_W$, which were still acceptable.

\begin{figure*}[!htbp]
    \centering
    \includegraphics[width=6.8in]{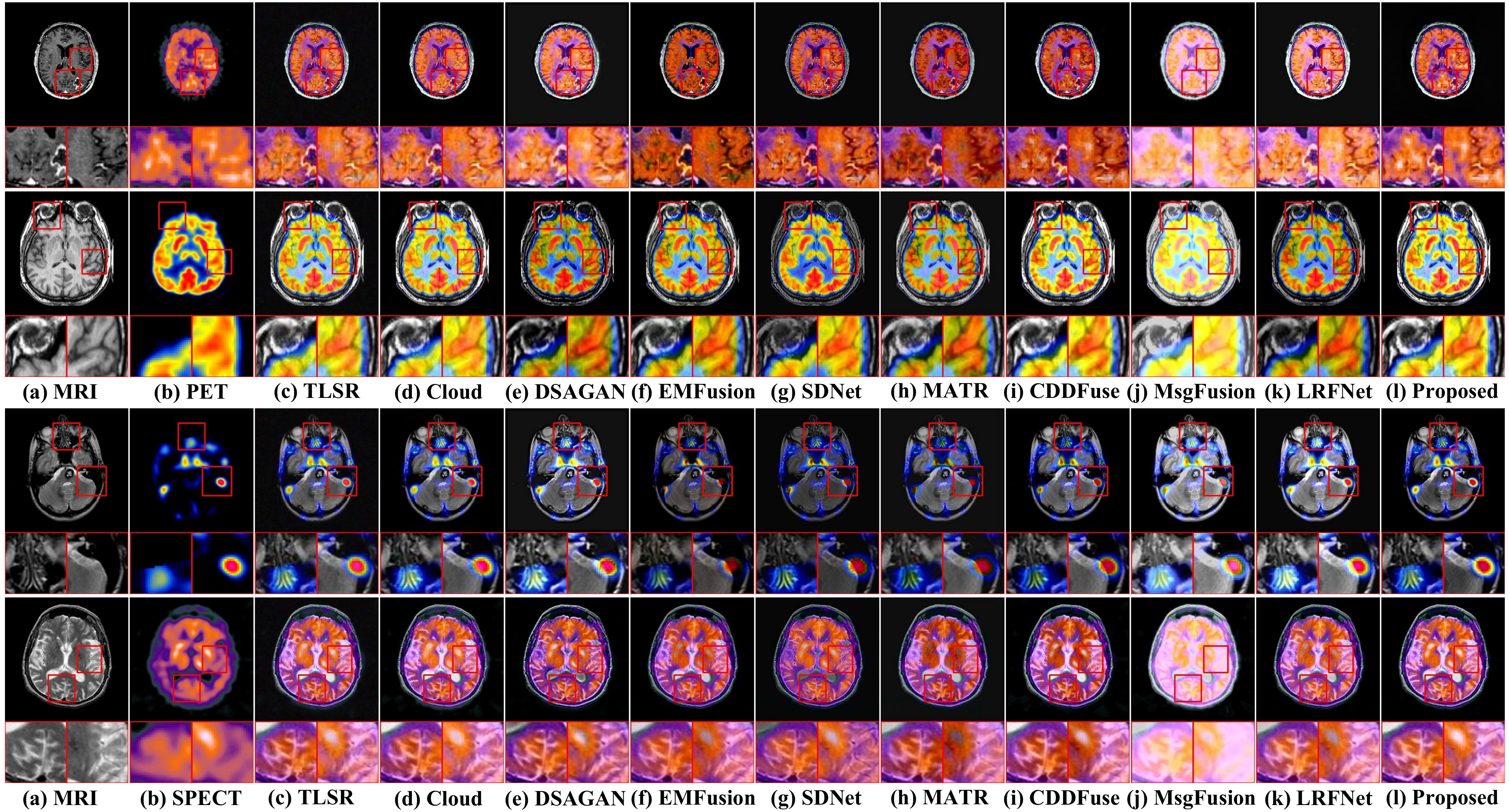}
    \caption{Visual comparison of DM-FNet with 9 SOAT methods for MRI-PET and MRI-SPECT image fusion. For a more intuitive comparison, the regions are enlarged as close-ups.}
    \label{Fig8}
\vspace{-0.2cm}
\end{figure*}

\begin{table*}[!htbp]
\renewcommand\arraystretch{1.3}
\begin{center}
\caption{Quantitative comparison of DM-FNet with 9 SOAT methods for MRI-PET and MRI-SPECT image fusion. The upper right corner displays the ranking of the method on this indicator. Red: best results, blue: second best.}
\label{tbl4-1}
    \begin{tabular}{lllllllllll}\hline
        Methods & SF ↑ & SD ↑ & AG ↑ & $Q_W$ ↑ & SCD ↑ & VIFF ↑ & $Q_{AB/F}$ ↑ & MSSSIM ↑ & FMI\_WT ↑ & Ranking ↓  \\ \hline
        TLSR & 23.944$^{6}$  & 64.750$^6$  & 7.441$^5$  & 0.871$^5$  & 1.105$^6$  & 0.498$^6$  & 0.565$^7$  & 0.906$^5$  & 0.166$^{10}$  & 56  \\
        Cloud & 25.782$^3$  & 69.332$^3$  & 7.633$^3$  & 0.888$^3$  & 1.378$^4$  & 0.536$^4$  & 0.664$^4$  & 0.915$^3$  & 0.278$^5$  & 32  \\
        DSAGAN & 25.090$^4$  & 58.775$^8$  & 7.541$^4$  & 0.778$^9$  & 1.101$^7$  & 0.461$^7$  & 0.514$^9$  & 0.889$^8$  & 0.208$^9$  & 65  \\
        EMFusion & 23.320$^7$  & 58.866$^7$  & 6.832$^7$  & 0.884$^4$  & 0.943$^8$  & 0.454$^8$  & \textcolor{blue}{0.678$^2$}  & 0.905$^6$  & \textcolor{red}{0.336$^1$}  & 50  \\
        SDNet & 22.425$^9$  & 49.715${^{10}}$  & 6.418$^{10}$  & 0.788$^8$  & 0.768$^9$  & 0.415$^9$  & 0.561$^8$  & 0.895$^7$  & 0.277$^6$  & 76  \\
        MATR & 21.564$^{10}$  & 53.360$^9$  & 6.555$^8$  & 0.847$^7$  & 0.198$^{10}$  & 0.362$^{10}$  & 0.635$^6$  & 0.839$^9$  & 0.218$^8$  & 77  \\
        CDDFuse & \textcolor{blue}{26.354$^2$}  & 67.662$^4$  & \textcolor{blue}{7.810$^2$}  & \textcolor{blue}{0.898$^2$}  & 1.222$^5$  & 0.512$^5$  & 0.677$^3$  & 0.907$^4$  & 0.300$^3$  & \textcolor{blue}{30}  \\
        MsgFusion & 22.724$^8$  & \textcolor{red}{91.261$^1$}  & 6.438$^9$  & 0.623$^{10}$  & \textcolor{red}{1.597$^1$}  & 0.558$^3$  & 0.394$^{10}$  & 0.810$^{10}$  & 0.245$^7$  & 59  \\
        LRFNet & 24.465$^5$  & 67.314$^5$  & 7.117$^6$  & 0.848$^6$  & 1.500$^3$  & \textcolor{blue}{0.570$^2$}  & 0.662$^5$  & \textcolor{blue}{0.921$^2$}  & \textcolor{blue}{0.320$^2$}  & 36  \\
        Proposed & \textcolor{red}{27.986$^1$}  & \textcolor{blue}{71.138$^2$}  & \textcolor{red}{8.234$^1$}  & \textcolor{red}{0.905$^1$}  & \textcolor{blue}{1.508$^2$}& \textcolor{red}{0.600$^1$}& \textcolor{red}{0.688$^1$}  & \textcolor{red}{0.945$^1$}& 0.278$^4$& \textcolor{red}{14}  \\ \hline
    \end{tabular}
\end{center}
\vspace{-0.2cm}
\end{table*}

{\bf{MRI and PET image fusion.}}
Fig. \ref{Fig8} presents the results of fusing MRI and PET images results. Two magnified regions are provided for each group for better comparison.
It was evident that EMFusion and MATR recovered high-brightness regions very poorly. In addition, the colors captured by EMFusion were distorted.
TLSR and the Cloud could preserve high-brightness regions of the PET images to some extent, but they sometimes did not capture these regions completely.
Although DSAGAN and SDNet were superior to the above four methods in terms of brightness recovery, they were prone to unstable blurring artifacts and darkened edges. This was related to DSAGAN's unstable training and SDNet's limited feature extraction capability.
MsgFusion was overly bright, resulting in the loss of many detailed features in the MRI image. Although LRFNet effectively extracts more information, it fails to recover the highlighted areas completely, and artifacts are present.
CDDFuse provided more satisfactory visual results, but slight color distortion still existed (see the first set of examples).
In contrast, DM-FNet emphasizes the texture and edge features of MRI images while accurately preserving the radioactive tracer distribution without artifacts in PET images, particularly in high-brightness regions.

{\bf{MRI and SPECT image fusion.}}
Fig. \ref{Fig8} shows the results of fusing the MRI and SPECT images.
Like MRI-PET image fusion, EMFusion, SDNet, MATR, and CDDFuse exhibited incomplete color recovery and poorly bright regions.
MsgFusion produced brighter images but resulted in blurriness and a loss of detail in the MRI images.
The PET features were overly emphasized in the DSAGAN method, which also exhibited artifacts.
Although LRFNet exhibited superior color recovery, it displayed diminished detail and darker highlighted regions.
TLSR and Cloud achieved more favorable visual results, whereas DM-FNet demonstrated superior color, detail information, and region recovery.

The nine objective evaluation metrics for the MRI-PET and MRI-SPECT image pairs are summarized in Table \ref{tbl4-1}.
The proposed method achieves optimal SF, $Q_W$, VIFF, and MSSSIM values, which indicates that it outperforms the other methods in terms of contrast, clarity, activity, and structural similarity.
Moreover, DM-FNet enhances the model’s sensitivity and capacity for capturing crucial details via its diffusion process and UNet’s multiscale structure. Consequently, DM-FNet is the most effective at recovering edges and textures, as demonstrated by its highest AG and $Q_{AB/F}$ values.
The high brightness of MsgFusion significantly affects its SD and SCD scores, resulting in DM-FNet receiving only the second-best scores on these two metrics.

\subsection{Computational efficiency analysis}
This section comprehensively evaluates the fusion efficiency via parameters, FLOPs, and testing time. Given the diverse data processing and training strategies employed in the compared methods, we defined the test time as the interval between invoking the model and obtaining the fusion results. The test time represents the average fusion time of 50 pairs of images.
\begin{table}[!t]
\renewcommand\arraystretch{1}
\begin{center}
\caption{Computational efficiency analysis of the model. Red: best results.}
\label{tbl5}
    \begin{tabular}{@{}lllllll@{}}\hline
        Methods & Device & Param (KB) & FLOPs (G) & Test Time/s  \\ \hline
        Cloud & CPU & - & - & 20.282  \\
        TLSR & CPU & - & - & 3.652  \\
        DSAGAN & GPU & 641 & 42.01 & 0.017  \\
        EMFusion & GPU & 150 & 19.64 & 0.035  \\
        SDNet & GPU & 67 & 8.81 & 0.006  \\
        CDDFuse & CPU & 1826 & 116.85 & 0.044  \\
        MATR & GPU & \textcolor{red}{11} & \textcolor{red}{1.95} & 0.055  \\
        MsgFusion & CPU & 2700 & 233.96 & 1.721  \\
        LRFNet & GPU & 55 & 4.08 & \textcolor{red}{0.004}  \\
        Proposed & GPU & 23774 & 252.36 & 0.329  \\ \hline
    \end{tabular}
\end{center}
\vspace{-0.5cm}
\end{table}

In the proposed method, the diffusion process is employed solely during the training phase to train the reconstruction-based UNet architecture and is not involved in the testing phase. This avoids the inefficiencies associated with the diffusion model's gradual denoising process.
As shown in Table \ref{tbl5}, the fusion time of the proposed method is 0.329 s, significantly outperforming traditional methods such as Cloud and TLSR, primarily owing to their time-consuming decomposition processes. In contrast, deep learning methods can leverage GPU acceleration, greatly improving efficiency.
Moreover, MsgFusion involves complex color conversion processes and lacks GPU acceleration, significantly impacting its fusion efficiency.
Compared with other deep learning methods, DM-FNet slightly increases the fusion time, but this trade-off is made in pursuit of superior fusion quality and more stable performance. By incorporating the diffusion process, DM-FNet achieves precise capture and retention of image details, which is particularly crucial in MIF.
Notably, medical image analysis typically occurs in the postprocessing stage, where real-time constraints are less stringent. Therefore, the high-quality fusion results and relatively efficient fusion time of DM-FNet make it a highly effective and reliable choice.

The model parameters are concentrated in the first stage, as mentioned in Section \ref{section4B}. Therefore, future research could further reduce the size of UNet in Stage I through model pruning or knowledge distillation to improve model efficiency. Specifically, analyzing the weight contribution of each layer and applying L1-norm pruning or structured pruning can eliminate redundant channels and filters. A lightweight single-stage student network can be trained using the original two-stage model as the teacher. Intermediate denoised results from the diffusion process serve as soft labels, with KL divergence loss constraining the student's feature distribution.

\subsection{Extension to PC-GFP image fusion}
\label{section4E}
\begin{figure*}[!htbp]
	\centering
	\begin{minipage}{1\linewidth}
		\centering
		\includegraphics[width=7in]{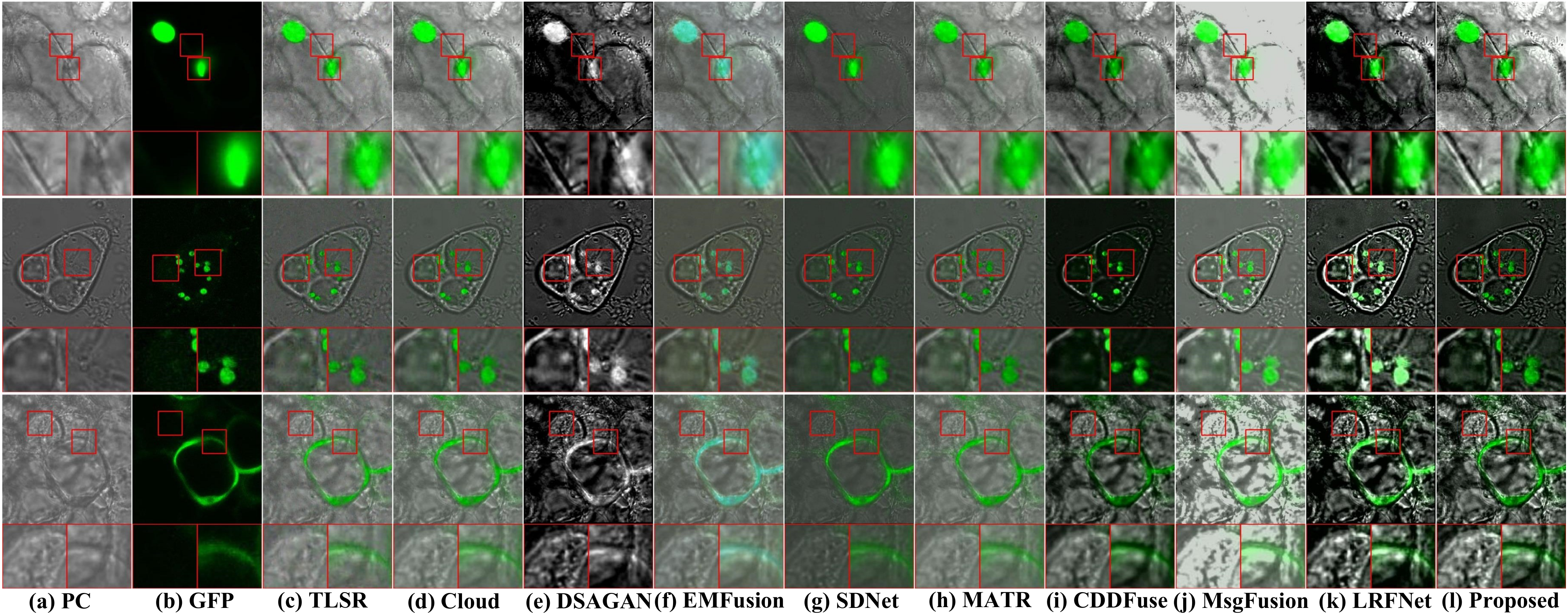}
		\caption{Visual comparison of DM-FNet with 9 SOAT methods for PC and GFP image fusion. For a more intuitive comparison, the regions are enlarged as close-ups.}
		\label{Fig11}
	\end{minipage}
	\begin{minipage}{1\linewidth}
		\centering
		\includegraphics[width=7in]{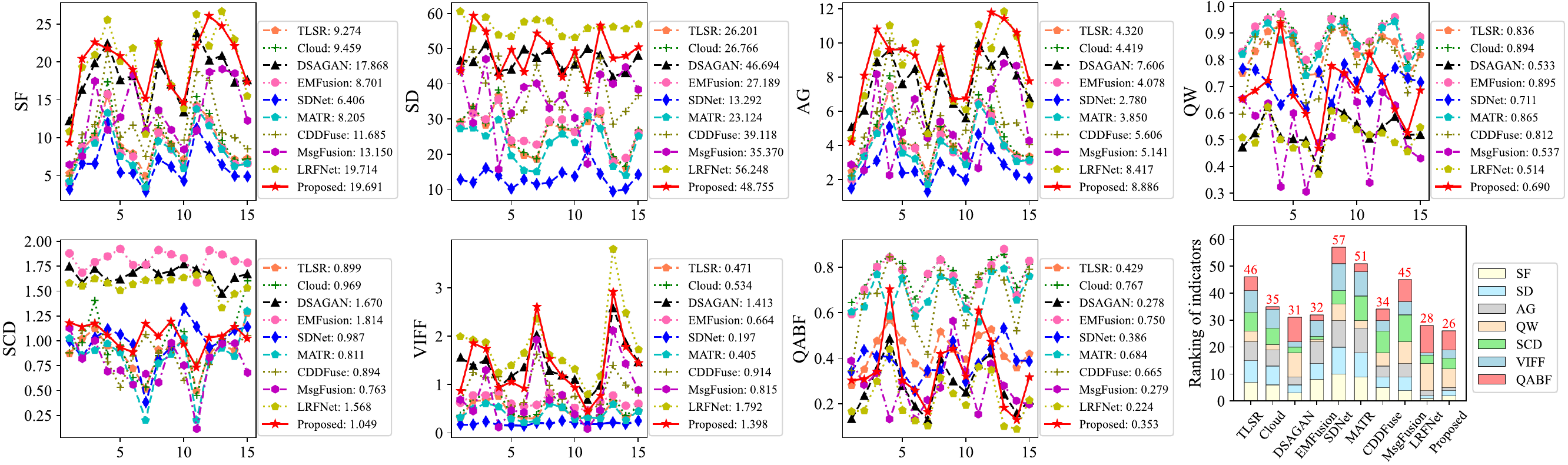}
		\caption{Quantitative comparison of DM-FNet with 9 SOAT methods for PC and GFP image fusion. The legend on the right displays the average scores for different methods.}
		\label{Fig12}
	\end{minipage}
\vspace{-0.5cm}
\end{figure*}
To verify the generalizability of DM-FNet, we fused PC and GFP images. The test set contained 15 image pairs. GFP images provide functional information related to the distribution of molecules in living cells, whereas PC images contain rich details of cellular structures, including the nucleus and mitochondria. GFP and PC fusion images can facilitate biological studies, such as gene expression and protein function analyses \cite{Xie2024FusionMambaDF}.

Fig. \ref{Fig11} shows the fusion results of the PC and GFP images.
EMFusion introduces color information to model training, resulting in significant color distortion for the new dataset.
TLSR and Cloud effectively retain the functional information of the GFP image but suffer from low intensity.
DSAGAN and LRFNet introduce significant noise while enhancing the details of the PC images.
MsgFusion could not extract sufficient structural information from the PC images.
CDDFuse achieves an optimal balance of visual quality, but it occasionally fails to capture PC image information fully (see the second set of examples).
DM-FNet preserves the complementary information from the source images and enhances the display of detailed information in the PC images.
Fig. \ref{Fig12} presents the results of the objective evaluation. The results demonstrate that identifying a method that outperforms existing methods in terms of most metrics is difficult. The proposed DM-FNet ranked highest in AG, second in terms of SD and SF, and first in terms of combined metrics, which indicates its good performance in the fusion task.

\subsection{Extension to infrared and visible image fusion}
\label{section4F}
\begin{figure*}[!tbp]
    \centering
    \includegraphics[width=7.1in]{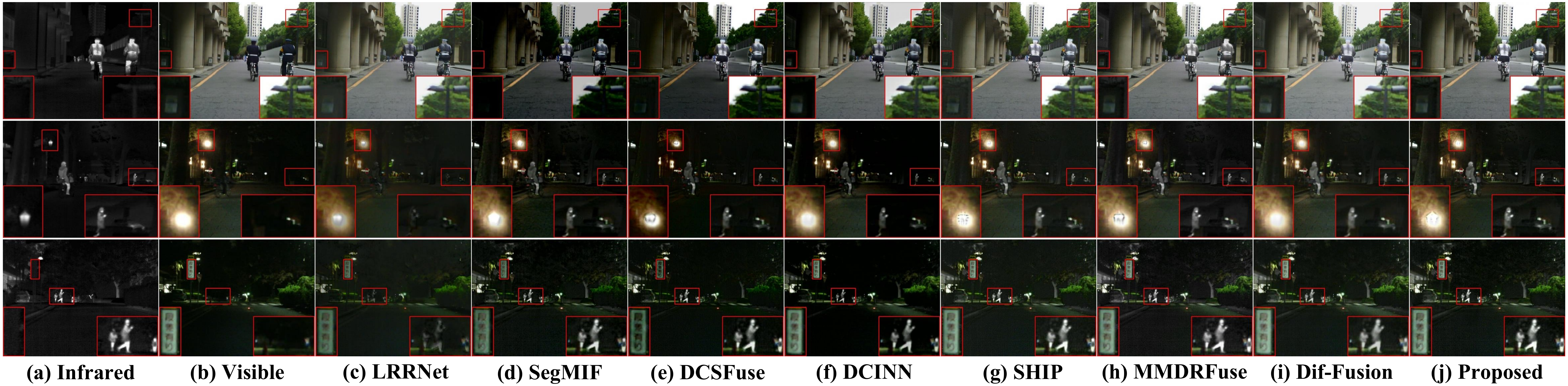}
    \caption{Visual comparison with 7 SOAT methods for IVIF on the MSRS dataset. For a more intuitive comparison, the regions are enlarged as close-ups. The three image pairs are 00537D, 00706N, and 00878N (from top to bottom).}
    \label{figivif}
\end{figure*}
\begin{table*}[!tbp]
\renewcommand\arraystretch{1.3}
\begin{center}
\caption{Quantitative comparison with 7 SOAT methods for IVIF. Red: best results, blue: second best.}
\label{tbivif}
    \begin{tabular}{lllllllllll}\hline
    Methods & SF ↑ & SD ↑ & AG ↑ & $Q_W$ ↑ & SCD ↑ & VIFF ↑ & $Q_{AB/F}$ ↑ & MSSSIM ↑ & FMI\_WT ↑ & Ranking ↓ \\ \hline
    LRRNet & 8.464$^{8}$  & 31.757$^{8}$  & 2.651$^{8}$  & 0.802$^{7}$  & 0.791$^{8}$  & 0.414$^{8}$  & 0.454$^{8}$  & 0.847$^{8}$  & 0.284$^{7}$   & 70 \\ 
    SegMIF & 10.417$^{7}$  & 35.963$^{7}$  & 3.309$^{7}$  & 0.776$^{8}$  & 1.511$^{5}$  & 0.645$^{6}$  & 0.550$^{6}$  & 0.933$^{6}$  & 0.323$^{4}$   & 56 \\ 
    DCSFuse & 11.541$^{4}$  & 38.745$^{6}$  & 3.572$^{4}$  & 0.856$^{6}$  & 1.400$^{7}$  & 0.640$^{7}$  & 0.546$^{7}$  & 0.921$^{7}$  & 0.301$^{5}$ & 53 \\ 
    DCINN & 10.467$^{5}$  & 40.112$^{5}$  & 3.334$^{6}$  & 0.882$^{5}$  & 1.478$^{6}$  & 0.706$^{4}$  & 0.568$^{5}$  & 0.953$^{5}$  & 0.337$^{3}$ & 44 \\ 
    SHIP & \textcolor{blue}{11.803$^{2}$}  & 41.131$^{3}$  & \textcolor{blue}{3.933$^{2}$}  & \textcolor{red}{0.926$^{1}$}  & 1.512$^{4}$  & 0.701$^{5}$  & \textcolor{red}{0.658$^{1}$}  & 0.962$^{4}$  & \textcolor{red}{0.348$^{1}$} & \textcolor{blue}{23} \\ 
    MMDRFuse & 10.426$^{6}$  & 40.248$^{4}$  & 3.494$^{5}$  & 0.901$^{4}$  & \textcolor{blue}{1.595$^{2}$}  & \textcolor{blue}{0.729$^{2}$}  & {0.590$^{3}$}  & {0.967$^{3}$}  & \textcolor{blue}{0.338$^{2}$} & 31 \\ 
    Dif-Fusion & {11.618$^{3}$}  & \textcolor{blue}{41.902$^{2}$}  & 3.890$^{3}$  & {0.914$^{3}$}  & 1.592$^{3}$  & 0.721$^{3}$  & 0.583$^{4}$  & \textcolor{blue}{0.967$^{2}$}  & 0.259$^{8}$ & 31 \\ 
    Proposed & \textcolor{red}{12.421$^{1}$} & \textcolor{red}{43.073$^{1}$} & \textcolor{red}{4.138$^{1}$} & \textcolor{blue}{0.925$^{2}$} & \textcolor{red}{1.654$^{1}$} & \textcolor{red}{0.781$^{1}$} & \textcolor{blue}{0.620$^{2}$} & \textcolor{red}{0.975$^{1}$} & 0.297$^{6}$ & \textcolor{red}{16} \\ \hline
    \end{tabular}
\end{center}
\vspace{-0.3cm}
\end{table*}

In this section, we fine-tune the second phase of the fusion network for expansion to IVIF.
The training and testing data are from the MSRS dataset, which contains 1081 pairs and 361 pairs of images. The experimental details and evaluation metrics are consistent with those used in MIF. 
The comparison methods include seven recent representative IVIF models: Dif-Fusion (2023)~\cite{Yue2023DifFusionTH}, LRRNet (2023)~\cite{Li2023LRRNetAN}, SegMIF (2023)~\cite{Liu2023SegMIF}, DCSFuse (2023)~\cite{cai2023correlation}, DCINN (2024)~\cite{wang2024general}, SHIP (2024)~\cite{10655996}, and MMDRFuse (2024)~\cite{Deng2024MMDRFuseDM}.

The fusion results of the SOAT method for three selected test images are shown in Fig. \ref{figivif}.
In the daytime scene, both SegMIF and DCSFuse tend to produce overly dark images, inadequately preserving the scene information in the visible image.
In the nighttime scene, LRRNet, DCSFuse, SHIP, and MMDRFuse exhibit an unnatural fusion of warm yellow lighting, leading to artifacts and reduced light spots. Furthermore, LRRNet fails to highlight key targets such as people and vehicles.
In contrast, DCINN, Dif-Fusion, and the proposed method demonstrate better fusion results for light sources and target objects (such as people, cars, and signs), avoiding artifacts and achieving superior scene information restoration.
The quantitative results (see Table \ref{tbivif}) indicate that the proposed method outperforms the other methods in six metrics, namely, SF, SD, AG, SCD, VIFF, and MSSSIM, further validating its SOAT performance.

\section{Conclusion}
This paper presents a fusion network, DM-FNet, that is based on the diffusion model. The network significantly improves the quality and information richness of the fused images by unifying the parameters to handle different types of medical image fusion tasks.
DM-FNet is a typical two-stage model.
In Stage I, a diffusion process trains a UNet for image reconstruction. During this process, detailed features and multilevel representations of the image are captured via a progressive denoising strategy, which ensures high-quality feature input for the fusion network.
In Stage II, the fusion network processes noisy images taken at different input steps. It incorporates three essential fusion modules: feature fusion of various noisy images, multimodal feature fusion, and multiscale feature fusion. This approach enables adaptive processing of medical images across different modalities, significantly enhancing the model's capacity for feature representation.

The experimental results demonstrate that the proposed method performs excellently for various medical image modalities. The fused images are well restored in brightness and color and exhibit high-definition texture and edges.
Moreover, the unfine-tuned model tested on the PC-GFP dataset and the fine-tuned fusion network evaluated on IVIF demonstrate strong generalizability. This success is largely attributed to the precise integration of multiple loss functions and the ability of the diffusion model to capture image details effectively.
Future work will focus on optimizing the algorithm to reduce the fusion time while ensuring high-quality results. Additionally, the model will be applied to downstream tasks, such as multimodal semantic segmentation and anomaly detection, to demonstrate its versatility across a broader range of applications. These efforts will help facilitate the wider adoption of this method in medical image processing.

{
\footnotesize
\bibliographystyle{unsrt}
\bibliography{myrefs}
}

\begin{IEEEbiography}[{\includegraphics[width=1in,height=1.25in,clip,keepaspectratio]{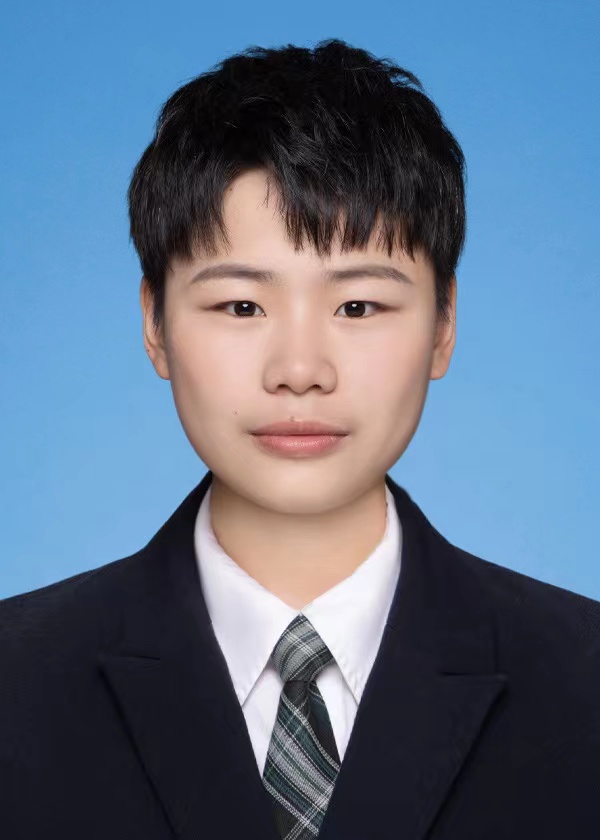}}]{Dan He}
received the B.S. degree from Hengyang Normal University, China, in 2020 and the M.S. degree from Chongqing Technology and Business University, China, in 2023, where she is currently pursuing the doctor's degree with the School of Computer Science and Technology, Chongqing University of Posts and Telecommunications. Her research interests include face anti-spoof, image fusion, and medical image processing.
\end{IEEEbiography}
\vspace{-1.5cm}

\begin{IEEEbiography}[{\includegraphics[width=1in,height=1.25in,clip,keepaspectratio]{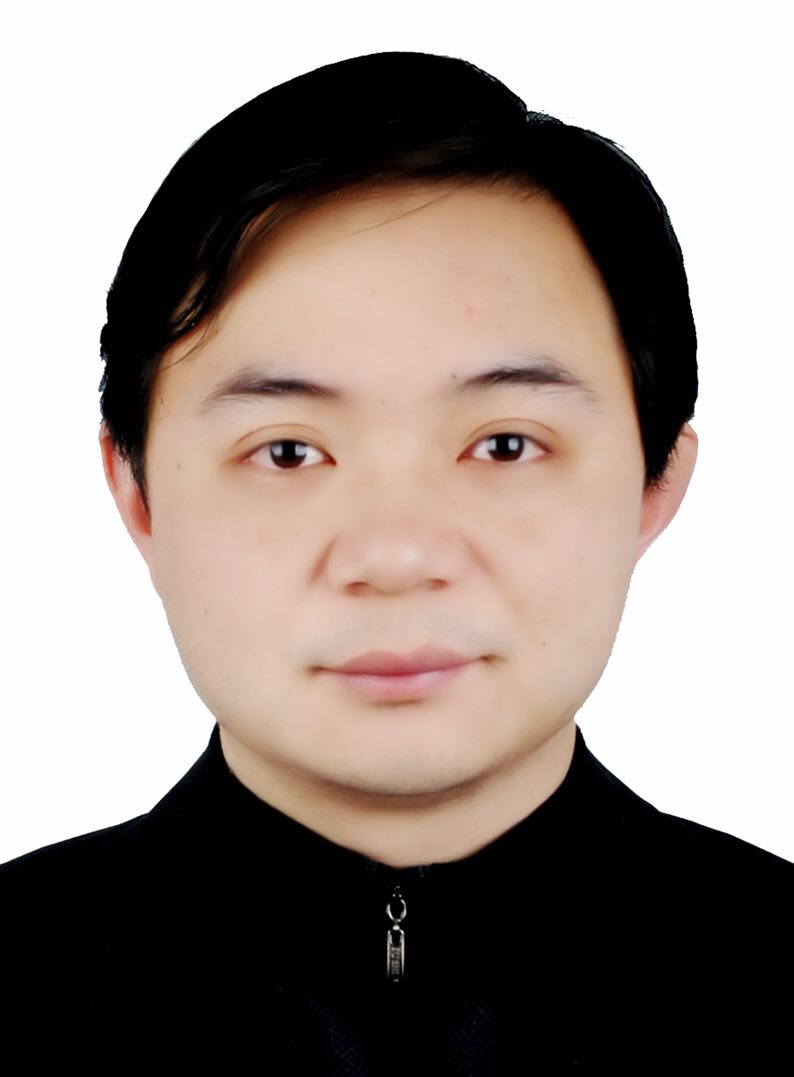}}]{Weisheng Li}
received the graduate degree from the School of Electronics and Mechanical Engineering, Xidian University, Xi’an, China, in July 1997, and the MS and PhD degrees in engineering and computer science from Xidian University in 2000 and 2004, respectively. He is currently a professor at the Chongqing University of Posts and Telecommunications and the director of the Chongqing Key Laboratory of Image Cognition. His research focuses on intelligent information processing and pattern recognition.
\end{IEEEbiography}
\vspace{-1.5cm}

\begin{IEEEbiography}[{\includegraphics[width=1in,height=1.25in,clip,keepaspectratio]{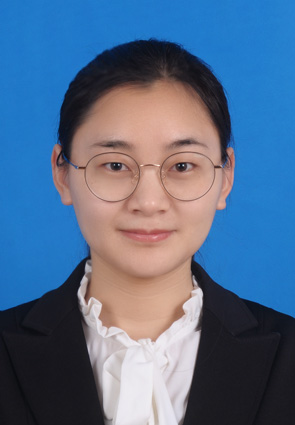}}]{Guofen Wang}
received the PhD degree from the School of Computer Science and Technology in Chongqing  University of Posts and Telecommunications, China, in 2023. She is currently a lecturer at Chongqing Normal University. Her research interests include medical image processing, image fusion, and optimization algorithms. 
\end{IEEEbiography}
\vspace{-1.5cm}

\begin{IEEEbiography}[{\includegraphics[width=1in,height=1.25in,clip,keepaspectratio]{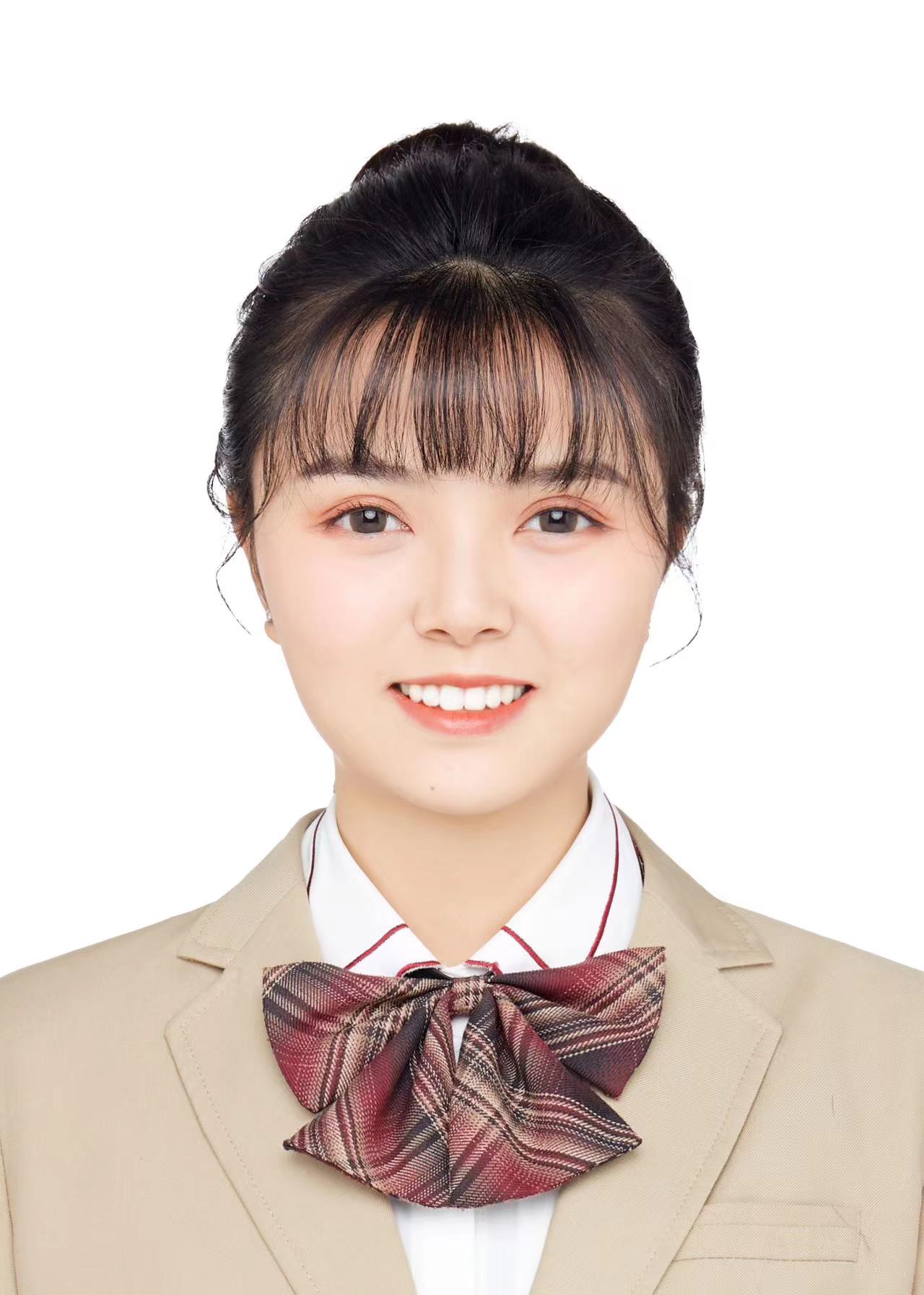}}]{Yuping Huang}
received the M.S. degree from Chongqing University of Posts and Telecommunications, China, in 2020, where she is currently pursuing the doctor's degree with the School of Computer Science and Technology, Chongqing University of Posts and Telecommunications. Her research interests include pattern recognition and medical image processing.
\end{IEEEbiography}
\vspace{-1.5cm}

\begin{IEEEbiography}[{\includegraphics[width=1in,height=1.25in,clip,keepaspectratio]{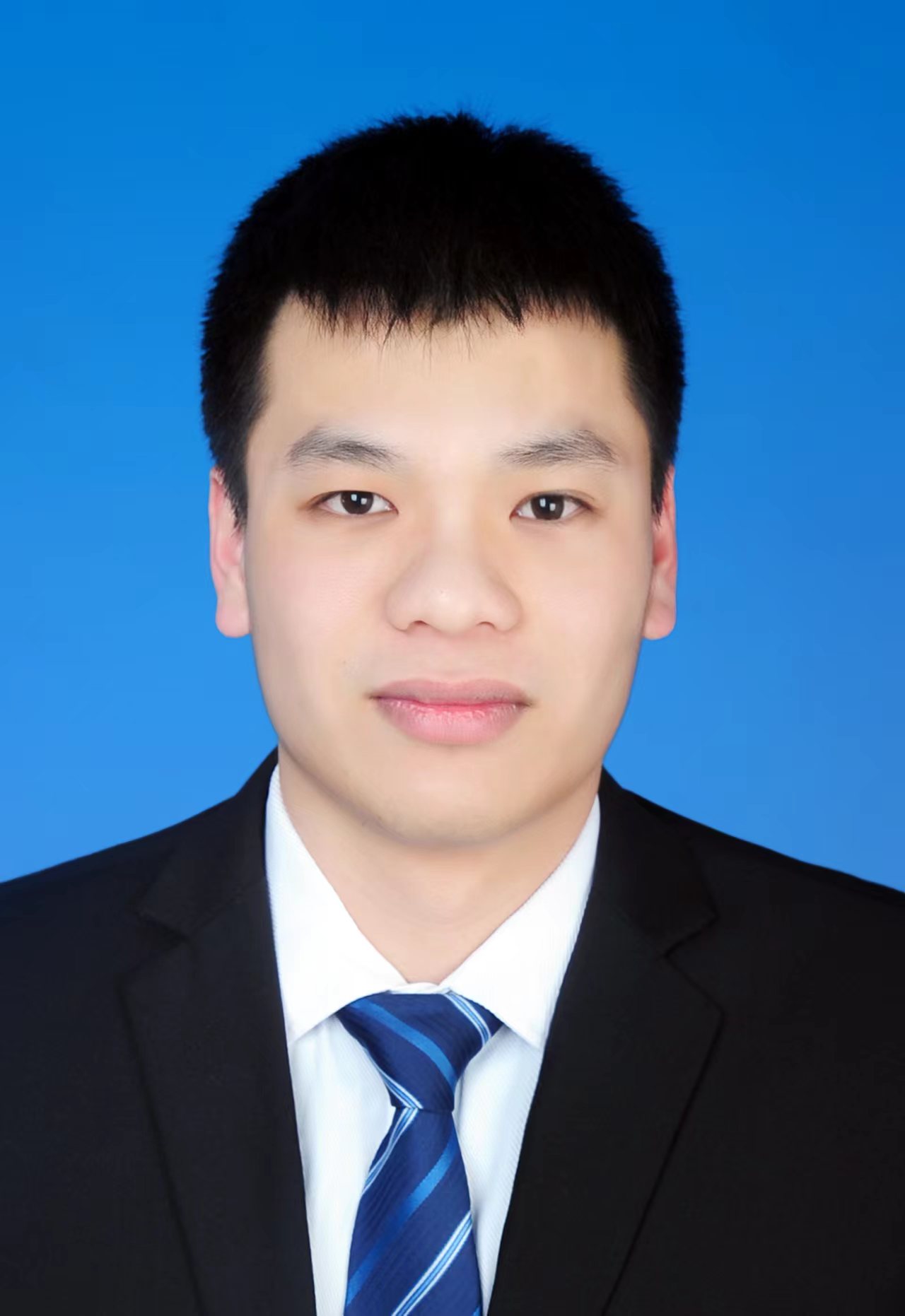}}]{Shiqiang Liu}
received the B.S. degree from the College of Computer Science and Technology, Chongqing University of Posts and Telecommunications, Chongqing, China in 2022. He is currently a Master's candidate in computer science and technology with the Chongqing Key Laboratory of Image Cognition, Chongqing University of Posts and Telecommunications, Chongqing, China. His research focuses on medical image fusion.
\end{IEEEbiography}


\end{document}